\documentclass[final,1p]{elsarticle}

\usepackage{amssymb}
\usepackage{xcolor}

\usepackage{lineno}

\usepackage{multirow}
\usepackage{amsmath}

\journal{Journal of Testing and Evaluation}

\begin{document}

\begin{frontmatter}



\title{Asphalt Concrete Characterization Using Digital Image Correlation: A Systematic Review of  Best Practices, Applications, and Future Vision}


\author[inst1]{Siqi Wang, Ph.D.}

\author[inst2]{Zehui Zhu, Ph.D.\corref{cor1}}

\affiliation[inst1]{organization={Department of Road Engineering, School of Transportation, Southeast University},
            addressline={Jiangning District}, 
            city={Nanjing}, 
            state={Jiangsu},
            postcode={211189},
            country={China}}

\affiliation[inst2]{organization={Department of Civil and Environmental Engineering, University of Illinois Urbana-Champaign},
            addressline={205 North Mathews Avenue}, 
            city={Urbana}, 
            state={IL},
            postcode={61801},
            country={United States}}

\author[inst1]{Tao Ma, Ph.D.}

\author[inst1]{Jianwei Fan, Ph.D.}

\cortext[cor1]{Corresponding Author}

\fntext[]{Accepted for publication in Journal of Testing and Evaluation. DOI: 10.1520/JTE20230485.}

\begin{abstract}
Digital Image Correlation (DIC) is an optical technique that measures displacement and strain by tracking pattern movement in a sequence of captured images during testing. DIC has gained recognition in asphalt pavement engineering since the early 2000s. However, users often perceive the DIC technique as an out-of-box tool and lack a thorough understanding of its operational and measurement principles. This article presents a state-of-art review of DIC as a crucial tool for laboratory testing of asphalt concrete (AC), primarily focusing on the widely utilized 2D-DIC and 3D-DIC techniques. To address frequently asked questions from users, the review thoroughly examines the optimal methods for preparing speckle patterns, configuring single-camera or dual-camera imaging systems, conducting DIC analyses, and exploring various applications. Furthermore, emerging DIC methodologies such as Digital Volume Correlation and deep-learning-based DIC are introduced, highlighting their potential for future applications in pavement engineering. The article also provides a comprehensive and reliable flowchart for implementing DIC in AC characterization. Finally, critical directions for future research are presented.
\end{abstract}




\end{frontmatter}


\section{Introduction}
Digital Image Correlation (DIC) is an optical-based method used for measuring displacements and strains in various materials. DIC functions by tracing patterns across a sequence of surface images of the specimen during testing \citep{pan2018digital}. This is achieved using subset-based matching, wherein gray value correspondences are extracted by identifying their resemblances \citep{chu1985applications,pan2011recent}. In 1982, \citet{peters1982digital} introduced the concept of extracting local surface deformations from single-camera images of planar specimens. A mathematical framework was proposed to convert digitized 2-D images into full-field displacement measurements, which is now known as two-dimensional digital image correlation (2D-DIC). Professor Sutton further contributed to the field by exploring implementation algorithms and applications \citep{sutton1983determination, peters1983application, he1984two}. However, it became evident in the mid-1980s that 2D-DIC was limited to flat specimens and single-plane deformations, which did not align with the requirements of most engineering studies. Consequently, the necessity for stereovision systems capable of capturing full-field three-dimensional displacement measurements on surfaces emerged. In the early 1990s, Professors Chao and Sutton developed the 3D-DIC stereovision system and successfully conducted experiments using it \citep{luo1993accurate, luo1994application}.

\begin{figure}[ht!]
    \centering
    \includegraphics[trim=0 0 0 0,clip,width=1\textwidth]{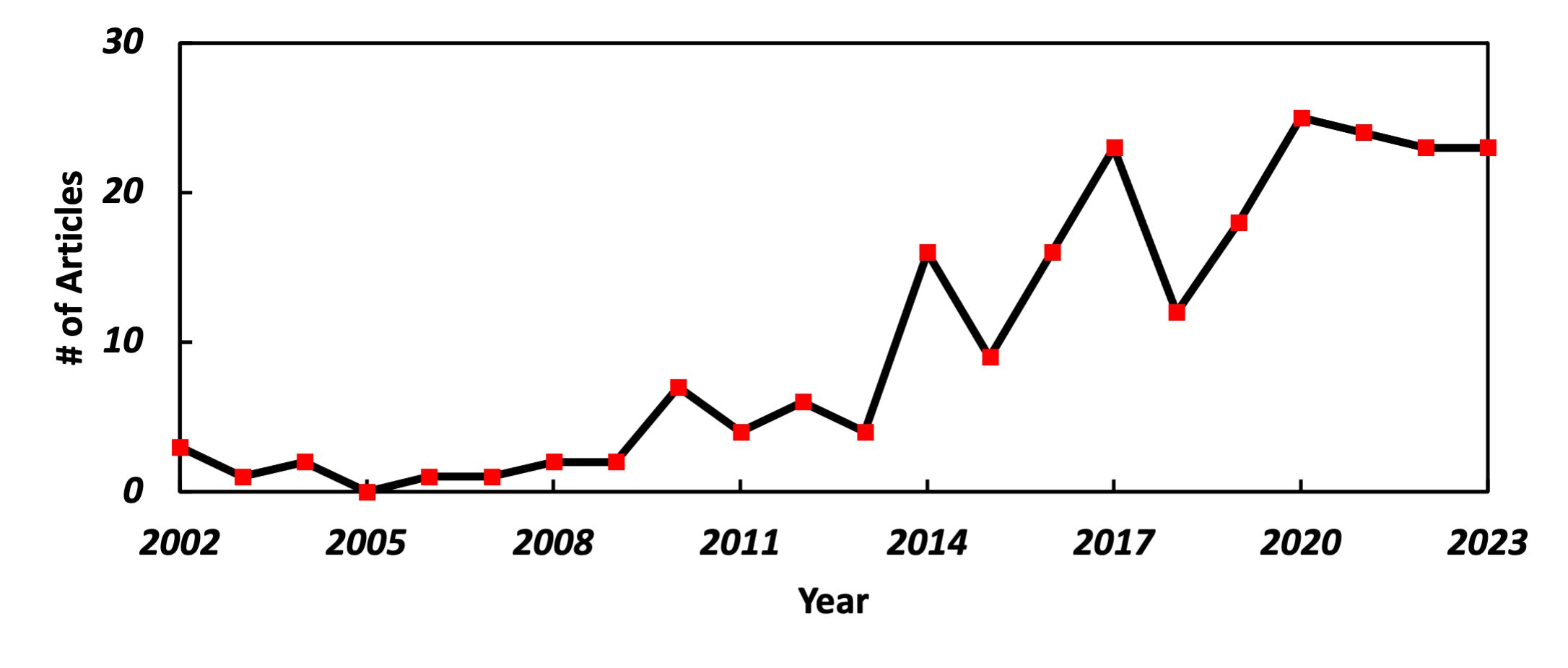}
    \caption{Number of articles published using DIC in AC characterization.}
    \label{fig:article}
\end{figure}

The accuracy and practicality of DIC have garnered attention within the asphalt pavement engineering community since the early 2000s. In 2002, \citet{seo2002application} pioneered the application of 2D-DIC to analyze the stress-strain behavior of the fracture process zone in monotonic and cyclic tests. Since then, DIC has become a prominent tool for evaluating the material properties of AC, validating experimental procedures, and verifying theoretical models \citep{chehab2007viscoelastoplastic,birgisson2008determination,birgisson2009optical,buttlar2014digital,safavizadeh2017utilizing,rivera2021illinois,zhu2023crack}. Figure \ref{fig:article} illustrates the number of scientific papers retrieved from the Web of Science (Science Citation Index Expanded) by using the search term \emph{digital image correlation asphalt}, covering the period from 2002 to 2023. The increasing number of published articles demonstrates the growing adoption of DIC in characterizing AC, particularly after 2014. It is important to note that 2D-DIC has received considerably more attention than 3D-DIC, primarily due to its simpler implementation (e.g., single-camera setup without the requirement of stereo camera calibration) \citep{zhu2023sift}. The dominance of 2D-DIC is evident as it accounts for over 95\% of the published articles in this field. Remarkably, the initial publication employing 3D-DIC emerged as recently as 2017 \citep{stewart2017comparison}.

Vendors commonly offer integrated software that enables users to obtain displacement and strain measurements. Consequently, users often perceive the DIC technique as an out-of-box tool and lack a thorough understanding of its operational and measurement principles. However, the accuracy of displacement and strain measurements obtained through DIC is significantly influenced by the specific implementation details employed \citep{pan2018digital}. Common inquiries from users encompass various aspects, such as the optimal preparation of specimen speckle and imaging system set up to attain the highest accuracy, approaches for assessing the precision of the DIC system, strategies for selecting user inputs in DIC analysis, understanding the underlying algorithms used by the DIC technique, and methods for post-processing and interpolating the measured displacement or strain maps to characterize AC. Hence, performing a thorough review of DIC can contribute to bolstering confidence in its usage and fostering standardization within the pavement engineering community.

This article provides a comprehensive and in-depth review of DIC as a crucial tool for laboratory testing of AC. The primary focus of this study centers around the widely employed 2D-DIC and 3D-DIC techniques. The article thoroughly examines the best practices pertaining to specimen speckle pattern preparation, the configuration of single-camera or dual-camera imaging systems, and the meticulous execution of DIC analyses. To enhance readers' understanding of the utility of DIC in their own work, the article documents experiences from over 100 publications spanning the past two decades, focusing on applying DIC-measured full-field displacement and strain maps for AC characterization. Furthermore, the article explores emerging DIC methodologies, including Digital Volume Correlation (DVC) and deep-learning-based DIC, which have not yet been adopted by the pavement engineering community but exhibit significant potential for future applications. Lastly, the article provides a flowchart intended to serve as a comprehensive and reliable reference for future DIC implementation in AC characterization.

\section{Specimen Preparation}

A crucial factor for accurate DIC measurements is the quality of the speckle pattern in use. The arrangement of speckles must possess specific attributes, as highlighted by Dong et al. in their comprehensive analysis \citep{dong2017review}. The term "high contrast" refers to the necessity of observing variations in grayscale intensities, resulting in significant intensity gradients among the speckles. The condition of "randomness" requires the absence of any repetitive or periodic elements within the speckle configuration. This absence is vital for achieving comprehensive displacement mapping across the entire field of view. "Isotropy" mandates that the speckle arrangement remains unbiased in all directions. Both the speckles and the spaces between them should maintain consistent dimensions across various orientations, as noted by \citet{reu2014all}. To prevent aliasing artifacts, it's advisable to use speckle granules sized around three to five pixels or slightly larger \citep{reu2015all}. Lastly, the concept of "stability" entails the firm adherence of the speckle pattern to the sample's surface. This adherence ensures that the pattern deforms coherently with the sample, even during significant translations and deformations. This stability should be upheld without causing noticeable changes in both geometric arrangement and grayscale characteristics.

The ongoing scientific discourse pertains to whether the natural texture of AC specimens meets the specified requirements. \citet{xing2017research} proposed that the natural texture of asphalt mixtures is suitable for DIC analysis, particularly in the case of AC with a small normal maximum aggregate size (NMAS). This perspective has received subsequent validation from various investigations, including \citet{guo2020investigation}. In contrast, \citet{yuan2020full} conducted a systematic comparison of DIC measurements using both artificially generated patterns and natural textures. Their findings indicated that the natural texture exhibited an error rate over three times greater than that of artificially generated speckle patterns. Moreover, the broader DIC research community tends to favor the generation of artificial speckle patterns to enhance measurement reliability and consistency \citep{pan2018digital,dong2017review}. Thus, the subsequent discussion will center on best practices for creating artificial speckle patterns.


\subsection{Artificial Speckle Pattern Fabrication}
citet{doll2015evaluation} examined the performance of three commonly employed methods in creating artificial speckle patterns for AC specimens. The evaluation included a comprehensive analysis of the intensity histograms, noise components, and measurement accuracy for simple motions involving rigid body translation and rotation. The three speckle pattern fabrication techniques assessed were as follows: 1) smoothing the sample surface using sandpaper and an airbrush, followed by the application of several light layers of white paint and a final layer of black paint \citep{seo2002application}; 2) applying a black paint layer to the surface and then generating the speckle pattern by spraying white paint on top of it; and 3) applying a thin layer of plaster to the specimen surface to fill in any holes (i.e., voids), followed by applying the speckle pattern on the plaster layer. 

The results showed that while there were no significant differences between the two patterns where the paint was applied directly to the asphalt, the pattern on plaster gave better results. However, using a coating can introduce other drawbacks, as the material at the surface may not behave the same way as the material underneath it, leading to inaccurate measurements. During fracture experiments, the authors observed that the plaster coating did not behave like the asphalt material, resulting in the peeling off of the plaster and inaccurate measurements. Hence, the authors suggest using the direct application of white and black paints without any coating, despite the resulting increase in measurement error caused by surface irregularities, as it enables accurate measurement of the material's displacements \citep{doll2015evaluation}.

\citet{lepage2017optimum} conducted an inquiry into whether superior results for DIC are achieved with white-on-black or black-on-white painted speckle patterns. Their findings identified the optimal speckle pattern composition as a white paint basecoat overlaid with black speckles. The study highlighted that black paint's greater concealing capacity and white paint's undertone due to Rayleigh scattering contributed to heightened contrast of the black speckles. Consequently, this increased contrast led to suggestions for reduced subset sizes, narrower correlation confidence intervals, higher mean intensity gradients, and ultimately more accurate displacement measurements (with a 24\% decrease in median normalized false displacement). As a result, the recommended painted speckle pattern entails a thin white paint basecoat, equal coverage for the basecoat and black speckles, and a speckle density of around 50\% \citep{lepage2017optimum}.

\begin{figure}[!ht]
  \centering
  \includegraphics[width=0.9\textwidth]{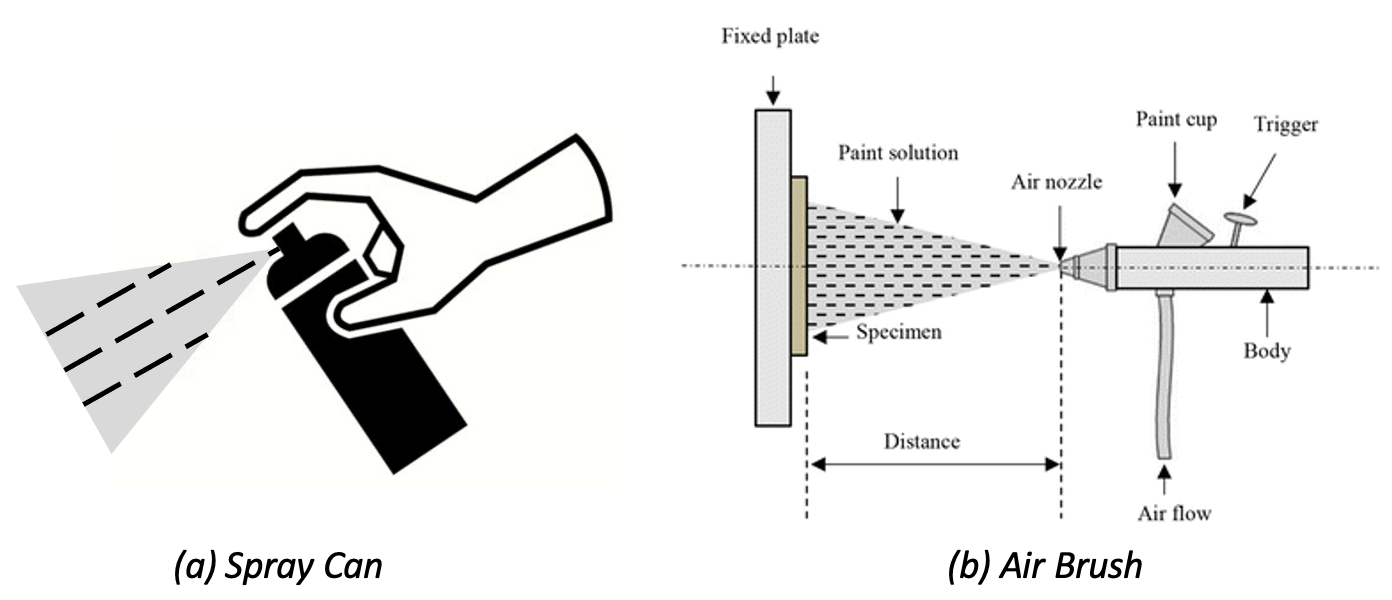}
  \caption{Schematic illustration of speckle pattern fabrication using a spray can or airbrush.}
  \label{fig:paint}
\end{figure}

Regarding the fabrication of speckle patterns, both spray bottles and airbrushes are commonly employed tools, as depicted in Figure \ref{fig:paint}. Nonetheless, there exists a variation in the size of resulting speckle granules, with spray bottle techniques generally yielding larger granules compared to airbrush methods. To exemplify, in their work, \citet{doll2017damage} utilized an airbrush to achieve an eight µm/pixel spatial resolution (i.e., the dimension of the pixel size representing the area covered on the specimen), allowing differentiation between strains in aggregate particles and the asphalt matrix regions between particles. Conversely, for assessing far-field strain and displacement fields, assuming homogeneous in-plane deformation, \citet{doll2017investigation} utilized a spray can, attaining an approximate spatial resolution of forty µm/pixel.

The dimensions of the nozzle, the gap between the nozzle and the specimen surface, air pressure, and solution viscosity are all pivotal factors that may impact the distribution of speckle sizes as well as the standard deviation of this distribution \citep{lionello2014practical}. Conducting preliminary trials is advisable to ensure that speckle granules measure around three to five pixels in size or marginally larger \citep{reu2015all}.

\subsection{Speckle Pattern Quality Assessment}
Different operators using various techniques to fabricate speckle patterns may result in different qualities, necessitating a quality assessment. Two categories of parameters are used to assess speckle patterns: local and global. Local parameters, such as subset entropy ($\delta$), sum of square of subset intensity gradients (SSSIG), and mean subset fluctuation ($S_f$), are designed to quantify individual subsets of the pattern and can assist with selecting the optimal subset sizes. On the other hand, global parameters, including mean intensity gradient (MIG) and Shannon entropy, quantify the entire speckle pattern. SSSIG and MIG are the most cited local and global metrics, respectively, owing to their solid theoretical foundations and straightforwardnformulations \citep{dong2017review, pan2008study, pan2010mean, neggers2016image}. 

To calculate SSSIG, Equation \ref{eqn:sssig} is used. The threshold of SSSIG can be determined based on the desired level of accuracy. Further information can be found in \citep{pan2008study}. SSSIG is frequently employed to assist in selecting the optimal subset size for DIC analysis, which will be discussed in a subsequent section. It should be noted, however, that SSSIG cannot distinguish between random and periodic speckle patterns as it only focuses on the local speckle pattern within an individual subset. 

\begin{equation}
  SSSIG=\sum_{i=1}^{N}\sum_{j=1}^{N}[f_{{x,y}}(\mathbf{x}_{ij})]
\label{eqn:sssig}
\end{equation}

where $f_{{x,y}}(\mathbf{x}_{ij})$ is the first derivative of the intensity gray value at pixel $\mathbf{x}_{ij}$ in $x$- or $y$- direction; $N$ is the subset size.

Equation \ref{eqn:mig} outlines the mathematical formula for calculating MIG. It is essential to note that a high value of MIG indicates a good speckle pattern. A recent study by \citet{zhu2023crack} found that a minimum MIG value of 25 produced a small displacement error. 

\begin{equation}
    MIG = \frac{\sum_{i=1}^W \sum_{j=1}^H |\nabla f(\mathbf{x}_{ij})|}{W \times H}
\label{eqn:mig}
\end{equation}

where $W$ and $H$ are image width and height in pixels, respectively; $|\nabla f(\mathbf{x}_{ij})|=\sqrt{f_x^2(\mathbf{x}_{ij})+f_y^2(\mathbf{x}_{ij})}$; $f_x(\mathbf{x}_{ij})$ and $f_y(\mathbf{x}_{ij})$ are the intensity derivatives at pixel ${x}_{ij}$ at the $x$- and $y$-direction, respectively. 

\section{2D-DIC}
The 2D-DIC method is a popular optical measurement technology due to its simple setup, minimal environmental prerequisites, and extensive sensitivity and resolution capabilities. It has become the primary DIC technology used in asphalt concrete characterization. However, limitations of the method include in-plane deformation measurement only, the need for a randomly distributed gray intensity on the object surface, and reliance on imaging system quality \citep{pan2009two}. This section will discuss best practices for 2D-DIC imaging system setup, algorithms, and applications in asphalt concrete characterization.

\subsection{Imaging System}
A frequently employed 2D DIC setup comprises a camera, illumination system, computer, and post-processing software. 2D-DIC requires high-quality images and control of imaging parameters for accurate measurements. 

\begin{figure}[!ht]
  \centering
  \includegraphics[width=0.55\textwidth]{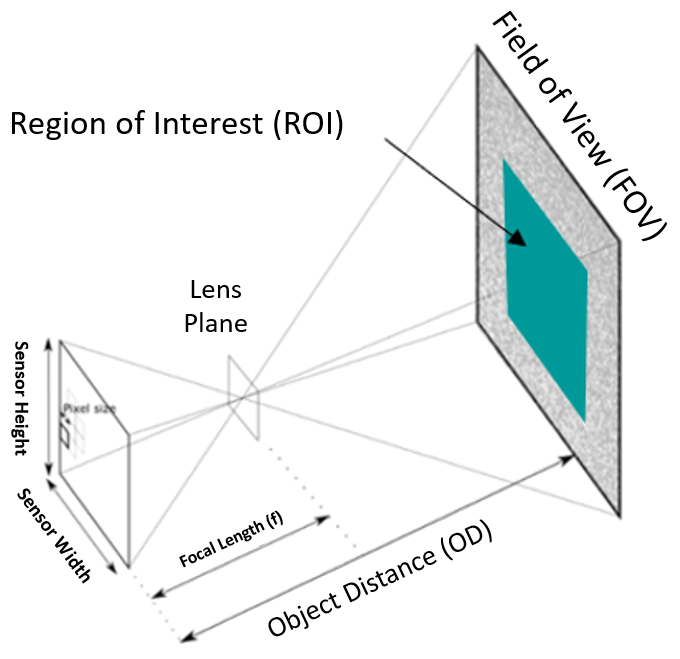}
  \caption{Pinhole camera model.}
  \label{fig:pinhole}
\end{figure}

First, to achieve optimal image quality and accurate measurement in 2D-DIC, it is essential to determine the optimal camera-object distance by fitting the Region-of-Interest (ROI) to the Field-of-View (FOV) as much as possible. The pinhole camera model (Figure \ref{fig:pinhole}) can be used to calculate the optimal distance between the camera and the object, given the selected lens.

For example, to measure the deformation of an AC specimen during a static loading test, with a specimen size of approximately 50$\times$50 mm, a standard 4 Megapixel camera with a sensor size of 2048$\times$2048 pixels and a pixel size of 5 $\mu$m is used. The actual dimension is slightly increased to ensure the object remains in view throughout the test. Using a 35 mm wide lens, the optimal distance required to fit the ROI within the FOV can be calculated (Equation \ref{eqn:od}):

\begin{equation}
  OD = \frac{w f}{S_w}+f = \frac{60 \times 35}{2048\times\frac{5}{1000}} + 35 \approx 240 mm
\label{eqn:od}
\end{equation}

where $OD$ is the optimal distance between the object and the sensor; $w$ is the dimension of the AC specimen, in this case, its width; $S_w$ is the corresponding dimension on the camera sensor, calculated as the number of pixels multiplied by the size of a pixel on the sensor, which is typically provided in the camera parameters; and $f$ is the focal length of the lens \citep{jones2018good}.

Then, for precise 2D-DIC measurements, it's essential for the camera's CCD sensor and the object's surface to be aligned in parallel. Additionally, any out-of-plane movement of the specimen during loading must be minimal, as emphasized by \citet{sutton2000advances}. Therefore, careful adjustment and positioning of the camera are essential, which can be challenging due to the lack of proper tools. 

A frequently employed method relies on the conventional computer vision camera calibration process \citep{zhang2000flexible}. Initially, at least ten calibration images is acquired using a standardized calibration plate. Subsequently, the calibration plate is held against the specimen to ensure its parallel alignment with the specimen. Next, a calibration procedure is executed, employing the final captured image and readily accessible tools, such as the MATLAB Camera Calibration Toolbox. Finally, iterative adjustments to the camera position are made and the aforementioned steps are iteratively repeated until an acceptable configuration is attained. It is essential to emphasize that this approach yields accuracy within a tolerance of approximately $\pm$2$^\circ$ \citep{wittevrongel2015evaluation}.

\begin{figure}[!ht]
  \centering
  \includegraphics[width=1\textwidth]{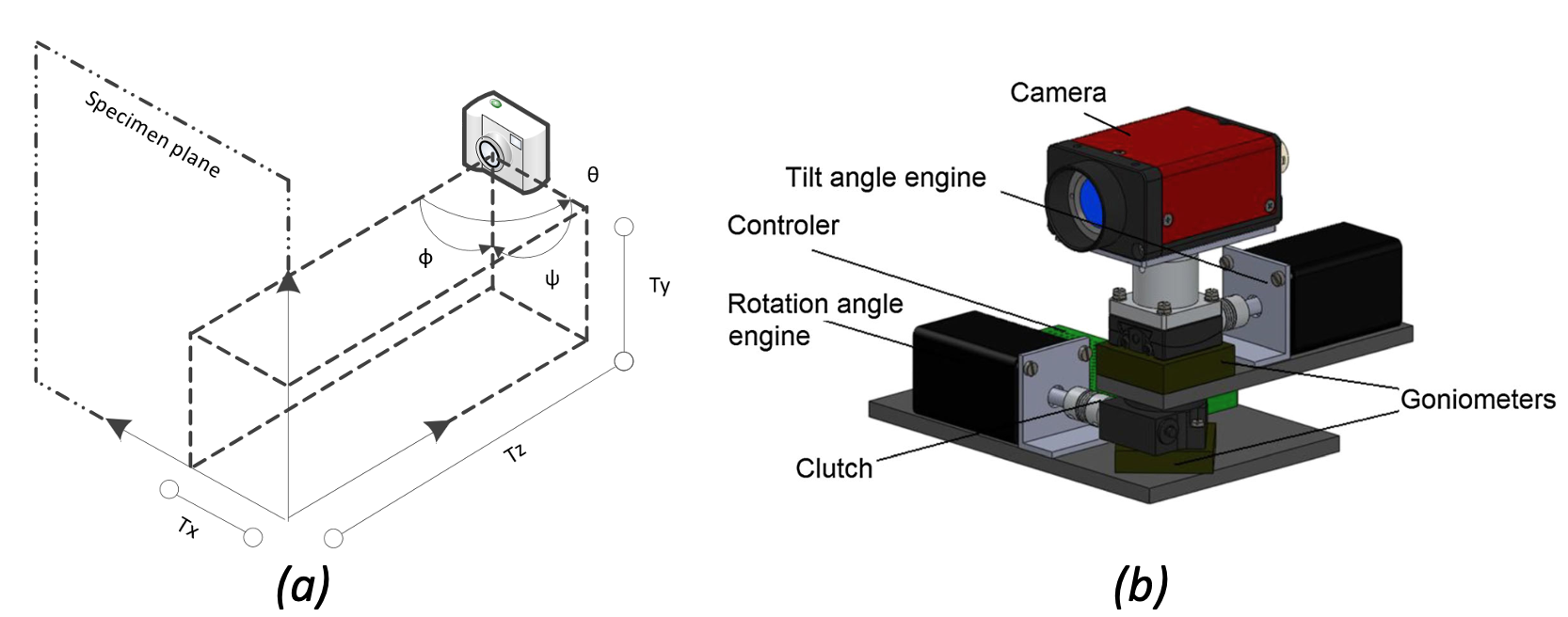}
  \caption{Mechanical camera positioning tool (a) camera's placement in relation to the specimen's surface; (b) diagram}.
  \label{fig:stage}
\end{figure}

\citet{wittevrongel2015evaluation} developed a high-precision rotation stage (Figure \ref{fig:stage}) using stepper engines to control the phi and theta angles, achieving precise control with 200 steps per revolution. The Psi angle is omitted, as rotational movements within the plane are permissible in 2D-DIC. The research findings demonstrate that the camera can be positioned accurately with a perpendicular precision of approximately $\pm$0.15$^\circ$. 

\begin{figure}[!ht]
  \centering
  \includegraphics[width=0.7\textwidth]{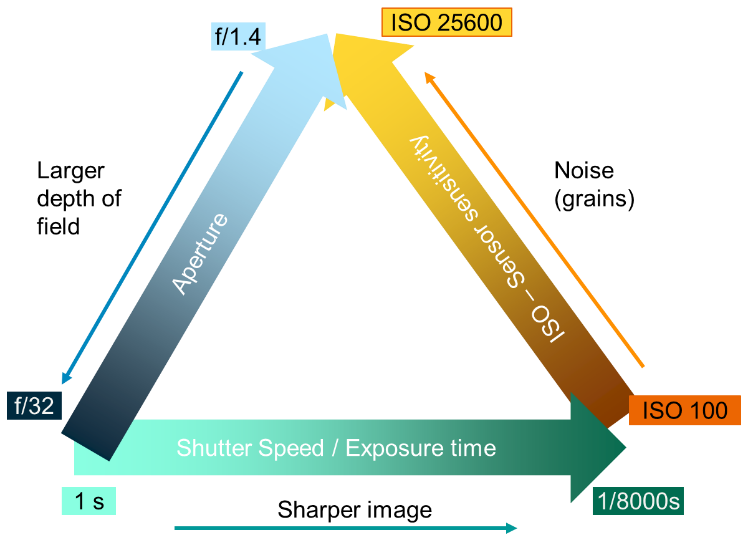}
  \caption{Exposure triangle.}
  \label{fig:exposure}
\end{figure}

Lastly, obtaining high-quality images involves adjusting the aperture, sensor sensitivity (ISO), and exposure time (shutter speed) to achieve sharp, well-illuminated images with minimal noise \citep{gruen2013calibration, jones2018good}. These three parameters are commonly referred to as the ``exposure triangle" and are determined by the properties of the lens and sensor. The relationship between these parameters is illustrated in Figure \ref{fig:exposure}. 

The ISO, or sensor sensitivity, is a property of the camera sensor. Increasing the ISO makes the sensor more sensitive to light and increases the image noise. To minimize noise, the ISO value should be kept at its factory default (usually ISO 100) and not changed.

The aperture, a property of the lens, controls the amount of light that enters the camera. A higher aperture allows more light to enter and narrows the depth of field, which may cause the background to be blurred. For flat objects like most AC specimens, positioned perpendicularly to the camera, narrow depths of field are generally sufficient. However, more complex structures with curved surfaces or different components may require a smaller aperture. 

Moreover, to ensure a sharp image during a real experiment, the exposure time must be shorter than the motion of the object being photographed. This is crucial for accurate measurements. Increasing sensor sensitivity is not recommended to avoid adding noise. Therefore, the solution is adding artificial light to brighten the scene.

\subsection{Algorithm}
\subsubsection{Fundamental Principles}
\begin{figure}[!ht]
  \centering
  \includegraphics[width=0.7\textwidth]{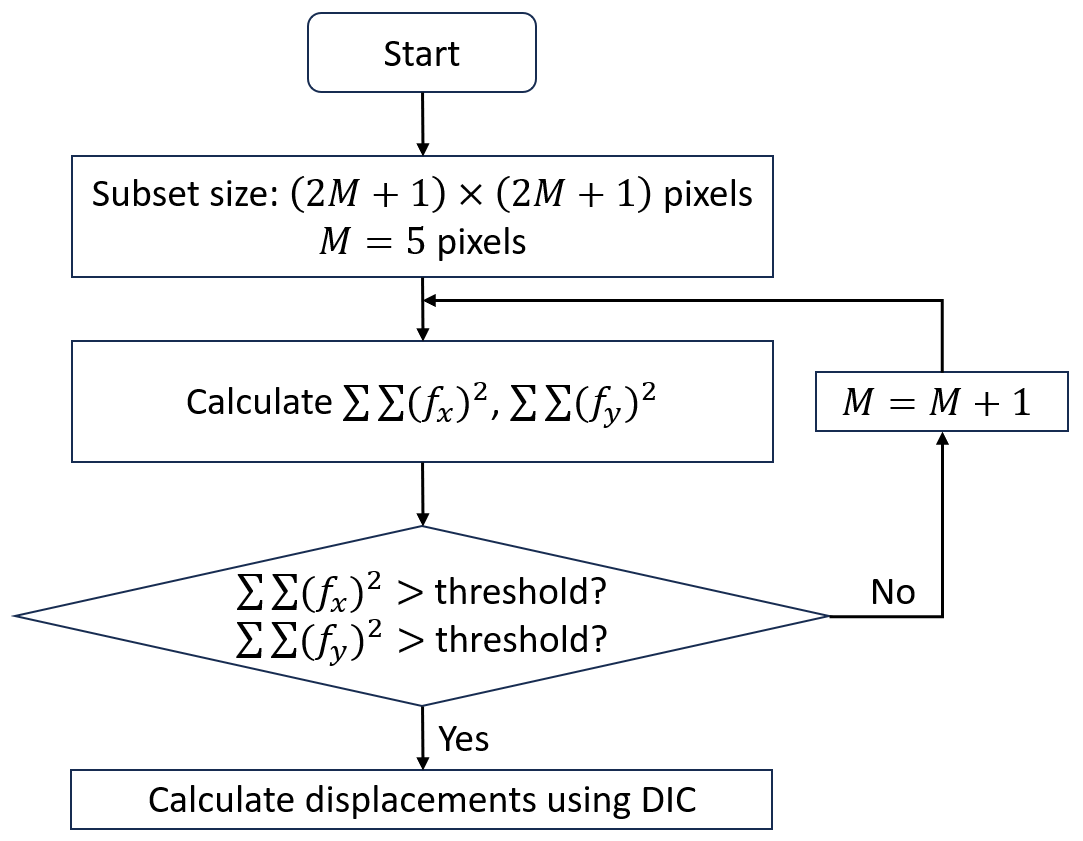}
  \caption{Flowchart of using SSSIG to select optimal subset size.}
  \label{fig:subset_size}
\end{figure}

\begin{figure}[ht!]
    \centering
    \includegraphics[width=0.95\textwidth]{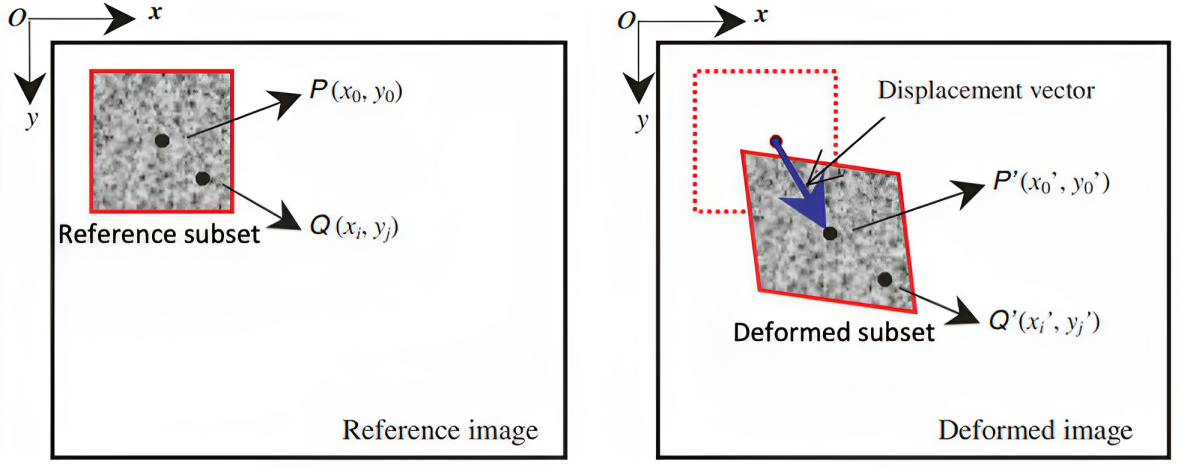}
    \caption{Area-based matching.}
    \label{fig:abm}
\end{figure}

DIC operates by monitoring pattern displacements in a series of images through subset-based matching, which identifies gray value correlations by comparing similarities. As depicted in Figure \ref{fig:abm}, when computing point $P$ displacements, a square subset of pixels $((2M+1) \times (2M+1))$ is selected from the reference image and matched with the deformed image. The subset size ($M$) is a crucial parameter in DIC analysis as it directly affects measurement accuracy. For dependable correlation analysis, the subset dimensions must balance distinct intensity patterns and accurate approximation of deformations using $1^{st}$ or $2^{nd}$-order subset shape functions. This balancing act calls for a compromise between employing larger or smaller subset sizes, as discussed by \citet{pan2008study}. SSSIG, which assesses speckle pattern quality, is commonly used to select the optimal subset size. Figure \ref{fig:subset_size} illustrates a flowchart outlining the process to identify the ideal subset size. This involves commencing with a smaller subset size and progressively enlarging it until the SSSIG surpasses the predefined threshold. The threshold can be determined based on the desired level of accuracy following the procedures presented in \citet{pan2008study}.

In the reference image, it's essential to designate an ROI that is subsequently partitioned into equidistant grids. The computation of displacements at each grid point facilitates the derivation of the displacement field.

The matching is attained by searching for a correlation coefficient extremum. Equation \ref{eqn:cost_fn} lists commonly used correlation criteria. Compared to cross-correlation (CC) and sum-of-squared differences (SSD), the zero-normalized cross-correlation (ZNCC) and zero-normalized sum-of-squared differences (ZNSSD) offer better performance against noise. They are less insensitive to lighting fluctuations (e.g., offset and linear scale) \citep{pan2009two}.

\begin{equation}
\resizebox{0.9\textwidth}{!}
{
$\begin{aligned}
    C_{SSD} &= \sum_{i=-M}^{M} \sum_{j=-M}^{M} [f(x_i,y_j) - g(x_i',y_j')]^2 \\ 
    C_{ZNSSD} &= \sum_{i=-M}^{M} \sum_{j=-M}^{M} [\frac{f(x_i,y_j)-f_m}{\sqrt{\sum_{i=-M}^{M} \sum_{j=-M}^{M} [f(x_i,y_j)-f_m]^2}} - \frac{g(x_i',y_j')-g_m}{\sqrt{\sum_{i=-M}^{M} \sum_{j=-M}^{M} [g(x_i',y_j')-g_m]^2}}]^2 \\
    C_{CC} &= \sum_{i=-M}^{M} \sum_{j=-M}^{M} f(x_i,y_j) g(x_i',y_j') \\
    C_{ZNCC} &= \frac{\sum_{i=-M}^{M} \sum_{j=-M}^{M}[f(x_i,y_j)-f_m]\times[g(x_i',y_j')-g_m]}{\sqrt{\sum_{i=-M}^{M} \sum_{j=-M}^{M} [f(x_i,y_j)-f_m]^2}\sqrt{\sum_{i=-M}^{M} \sum_{j=-M}^{M} [g(x_i',y_j')-g_m]^2}}\\
\end{aligned}$
}
\label{eqn:cost_fn}
\end{equation}

where $f(x_i,y_j)$ is gray value at $(x_i, y_j)$ in the reference subset. $g(x_i',y_j')$ is gray value at $(x_i', y_j')$ in the deformed subset. $f_m$ and $g_m$ are mean gray values of the reference and deformed subset, respectively.

A correlation coefficient or ZNSSD cost approaching zero indicates a favorable match. ZNCC is directly related to ZNSSD. Equation \ref{eqn:cost_relation} indicates that a $C_{ZNCC}$ value of 1 signifies a perfect match, while a value of 0 denotes no correlation.

\begin{equation}
    C_{ZNCC} = 1-0.5C_{ZNSSD}
\label{eqn:cost_relation}
\end{equation}

In Equation \ref{eqn:cost_fn}, the reference point $(x_i, y_j)$ is associated with the deformed point $g(x_i', y_j')$ through a mapping function. This function can be either $1^{st}$-order (as in Equation \ref{eqn:displ_first_fn}) or $2^{nd}$-order (as in Equation \ref{eqn:displ_second_fn}). The $2^{nd}$-order function has the ability to approximate more intricate displacements compared to the $1^{st}$-order one.

\begin{equation}
    \begin{bmatrix} x_i' \\ y_j' \end{bmatrix} = \begin{bmatrix} x_0 \\ y_0 \end{bmatrix}+\begin{bmatrix} 1+u_x & u_y & u \\ v_x & 1+v_y &v \end{bmatrix} \begin{bmatrix} \Delta x \\\Delta y \\ 1 \end{bmatrix}
\label{eqn:displ_first_fn}
\end{equation}

\begin{equation}
    \begin{bmatrix} x_i' \\ y_j' \end{bmatrix} = \begin{bmatrix} x_0 \\ y_0 \end{bmatrix}+\begin{bmatrix} 1+u_x & u_y & \frac{1}{2}u_{xx} & \frac{1}{2}u_{yy}& u_{xy}& u\\ v_x & 1+v_y & \frac{1}{2}v_{xx} & \frac{1}{2}v_{yy}& v_{xy}& v \end{bmatrix} \begin{bmatrix} \Delta x \\\Delta y \\ \Delta x^2 \\\Delta y^2 \\\Delta x \Delta y \\ 1 \end{bmatrix}
\label{eqn:displ_second_fn}
\end{equation}

Here, $u$ and $v$ represent the horizontal and vertical displacement components for the subset center $(x_0, y_0)$, respectively. The quantities $\Delta x = x_i - x_0$ and $\Delta y = y_j - y_0$ are defined. Additionally, $u_x$, $u_y$, $v_x$, and $v_y$ signify the components of $1^{st}$-order displacement gradients. Furthermore, $u_{xx}$, $u_{yy}$, $u_{xy}$, $v_{xx}$, $v_{yy}$, and $v_{xy}$ denote the components of $2^{nd}$-order displacement gradients. This paper employs the notation $\mathbf{p}$ to represent the desired displacement vector, which consists of either 6 or 12 unknown parameters.

Given the aforementioned definitions, it is evident that the computation of $\mathbf{p}$ entails an optimization task involving a user-defined cost function, such as Equation \ref{eqn:cost_fn} and Equation \ref{eqn:cost_relation}. DIC employs the Newton–Raphson (NR) iterative approach for optimization, as outlined in Equation \ref{eqn:NR}.

\begin{equation}
    \mathbf{p} = \mathbf{p}_0-\frac{\nabla C(\mathbf{p}_0)}{\nabla \nabla C(\mathbf{p}_0)}
\label{eqn:NR}
\end{equation}

Here, $\mathbf{p}_0$ represents the initial estimation of the displacement vector, while $\mathbf{p}$ denotes the subsequent iterative solution. The symbol $\nabla C(\mathbf{p}_0)$ corresponds to the $1^{st}$-order derivatives of the cost function, and the term $\nabla \nabla C(\mathbf{p}_0)$ refers to the Hessian matrix \citep{pan2009two}.

\subsubsection{RG-DIC}
The preceding section exclusively delineated the process of computing $\mathbf{p}$ for an individual point. To achieve full-field displacement measurement, the reliability-guided digital image correlation (RG-DIC) technique is widely employed \citep{pan2009reliability}. This approach has been integrated into open-source solutions such as Ncorr \citep{pan2009reliability, blaber2015ncorr}.

The process commences with the determination of an initial displacement vector estimate for a user-defined reference point. To achieve this, one might employ normalized cross-correlation or the scale-invariant feature transform (SIFT) technique for an informed initial estimation. Following this, the algorithm computes $\mathbf{p}_{seed}$ and its associated correlation coefficient. Subsequently, the algorithm computes the displacement vectors and correlation coefficients for the four adjacent points of the seed point, utilizing $\mathbf{p}_{seed}$ as the initial approximation. These computed correlation coefficients are incorporated into a priority queue. The subsequent step involves extracting the highest-correlation point from the queue and utilizing its corresponding $\mathbf{p}$ as the starting point to compute displacements for its neighboring points, if they are yet to be computed. This process iterates until the priority queue becomes empty, signifying the calculation of all points within the ROI.

Due to its incorporation of correlation analysis focused on points with the highest correlation, the RG-DIC approach exhibits resilience to minor image discontinuities. This attribute enhances its efficacy in analyzing images of AC specimen surfaces, which often exhibit irregularities arising from factors like air voids and diverse aggregate orientation.

The RG-DIC method may encounter challenges when dealing with substantial discontinuities, such as cracks, within the deformed image. This issue commonly arises during the analysis of deformed images obtained during the post-peak load stage of AC testing. To mitigate the decorrelation problem, \citet{zhu2023crack} introduced the multi-seed incremental approach. In this context, ``multi-seed" entails manually placing seed points on all partitions artificially created by the significant discontinuities, while ``incremental analysis" involves using an intermediate deformed image as an updated reference image if the deformed image exhibits severe decorrelation with the original reference image. When correctly implemented, the multi-seed incremental RG-DIC analysis consistently attains high accuracy, even in the presence of substantial discontinuities (such as cracks) in the deformed image.

\subsubsection{Compute of Strains}
In the realm of AC characterization, complete strain distributions frequently hold greater significance and desirability compared to displacement fields. However, strains are more challenging to resolve than displacement fields due to their sensitivity to noise caused by differentiation \citep{pan2009two, blaber2015ncorr}. Thus, it is necessary to smooth displacement fields before calculating strain fields. 

An illustration of this concept is the strain window technique introduced by \citet{pan2009digital}, wherein displacement gradients and Green-Lagrangian strains are computed through a least squares plane fit applied to a subset of displacement information. Subsequently, the algorithm resolves an excessive system of equations to ascertain the strains, offering flexibility in adjusting the size of the subset window. More details can be found elsewhere \citep{blaber2015ncorr, pan2009digital, eberly2000least}. 

Upon the parameter solution, they are employed for the computation of $e_{xx}$, $e_{xy}$, and $e_{y}$ as per Equation \ref{eqn:strain}. This procedure is then extended across the entire displacement field to derive the corresponding strain field.

\begin{equation}
\begin{aligned}
    e_{xx} &= \frac{1}{2}(2\frac{\partial u}{\partial x} + (\frac{\partial u}{\partial x})^2 + (\frac{\partial v}{\partial x})^2) \\ 
    e_{xy} &= \frac{1}{2}(\frac{\partial u}{\partial y} + \frac{\partial v}{\partial x} + \frac{\partial u}{\partial x}\frac{\partial u}{\partial y} + \frac{\partial v}{\partial x}\frac{\partial v}{\partial y}) \\ 
    e_{yy} &= \frac{1}{2}(2\frac{\partial v}{\partial y} + (\frac{\partial u}{\partial y})^2 + (\frac{\partial v}{\partial y})^2) \\
\end{aligned}
\label{eqn:strain}
\end{equation}

It is vital to emphasize that achieving reliable and precise full-field strain estimation necessitates the careful selection of an appropriate local strain calculation window size. For uniform deformation, a larger window size for strain calculation is preferable. However, when dealing with non-uniform deformation, the choice of strain calculation window size should be deliberate, considering the interplay between strain accuracy and smoothness. A small window might not effectively mitigate displacement noise, whereas an overly large window might yield an impractical linear deformation approximation within the strain calculation window \citep{pan2009two, pan2009digital, pan2007full}.

\subsubsection{Software}
The effective implementation of algorithms is crucial for 2D-DIC analysis. Table \ref{tab:2d} presents a compilation of presently accessible non-commercial 2D-DIC software. It is essential to note that this list exclusively comprises software with supporting peer-reviewed research papers, and there may be other available choices.

\begin{table}[!ht]
\centering
\resizebox{\textwidth}{!}{\begin{tabular}{c|c|c|c|c|c|c|c}
\hline
\hline
Software & Authors  & First Release                & User Interface    & Open Source & Free & Language & Citations (Oct 2023) \\
\hline
NCorr    & \citet{blaber2015ncorr} & 2015 & Yes & Yes         & Yes  & Matlab & 1,629 \\
\hline
DICe     & \citet{turner2015digital} & 2015      & Yes               & Yes         & Yes  & C++ & 12 \\
\hline
$\mu$DIC    & \citet{olufsen2020mudic} & 2019 & No & Yes         & Yes  & Python & 36 \\
\hline
OpenCorr    & \citet{jiang2023opencorr} & 2021 & No & Yes         & Yes  & C++ & 6\\
\hline
iCorrVision-2D   & \citet{de2022icorrvision} & 2022 & Yes & Yes         & Yes  & Python & 7 \\
\hline
\hline
\end{tabular}}
\caption{2D-DIC Software.}
\label{tab:2d}
\end{table}

\subsection{Applications}
\label{sec:2dapplications}
The 2D-DIC technique is extensively used in the asphalt pavement field to assess the material properties of AC. It is recognized as a practical and robust method for quantifying the deformation of AC specimens in diverse laboratory testing environments. A review of over 100 academic papers, with a specific emphasis on publications from 2017 onwards, revealed that the semi-circular bending (SCB) test, indirect tensile test (IDT), three-point bending (3PB) test, and four-point bending (4PB) test were the most frequently observed applications of the 2D-DIC technique. A limited number of studies employed the single-edge notched bending (SENB) test, disk-shaped compact tension (DCT) test, direct tension test (DTT), and triaxial repeated creep test methodologies.

Table \ref{tab:2d_application} summarizes the application of 2D-DIC in reviewed academic papers. Broadly, the applications of 2D-DIC can be classified into three categories: direct application of 2D-DIC-generated displacement or strain maps, derivation of mechanistic parameters for AC material properties, and tracking of crack propagation or damage evolution. The following sections provide detailed discussions of these applications.

\begin{table}[!ht]
\centering
\resizebox{\textwidth}{!}{\begin{tabular}{c|c|c|c}
\hline
\hline
Type                      & Test                   & Applications           & Articles \\ \cline{1-4}
\multirow{8}{*}{Fracture} & \multirow{3}{*}{SCB}   & Strain \& Displacement &     \citep{radeef2022influence,wang2022multiscale,wu2022influence,cui2022effect,stewart2017comparison}     \\ \cline{3-4}
                          &                        & Mechanistic Parameters    &    \citep{radeef2023fracture,hu2022use,pei2021effects,al2022cracking,rivera2021illinois,zhu2020crack,doll2017damage,doll2017investigation}      \\ \cline{3-4}
                          &                        & Crack Propagation      &    \citep{radeef2023fracture,kong2023evaluating,wu2022performance,al2022cracking,hu2022use,asghar2022evaluation,pei2021effects,zhu2020digital,zhou2017cracking}      \\ \cline{2-4}
                          & \multirow{2}{*}{3PB}   & Strain \& Displacement &     \citep{wu2022evaluation,hill2017inverse}     \\ \cline{3-4}
                          &                        & Crack Propagation      &     \citep{lin2023comprehensive,wang2022study,cullen2021assessing}     \\ \cline{2-4}
                          & \multirow{2}{*}{SENB}  & Strain \& Displacement &    \citep{wang2021research}      \\ \cline{3-4}
                          &                        & Crack Propagation      &    \citep{wang2020macro}      \\ \cline{2-4}
                          & DCT                    & Strain \& Displacement &      \citep{wang2018microwave,stewart2017comparison,hill2017inverse}    \\ 
\hline
\multirow{5}{*}{Fatigue}  & \multirow{2}{*}{4PB}   & Mechanistic Parameters    &     \citep{pedraza19fracture}     \\ \cline{3-4}
                          &                        & Crack Propagation      &     \citep{safavizadeh2022interface,pedraza19fracture,sudarsanan2019digital,kumar2017use,safavizadeh2017dic,wargo2017comparing,safavizadeh2015investigating}     \\ \cline{2-4}
                          & \multirow{2}{*}{SCB}   & Strain \& Displacement &     \citep{radeef2022linear,wang2022multiscale,radeef2021characterisation,yang2021feasibility,jiang2019evaluation,jiang2018fatigue,zhang2018characterization}     \\ \cline{3-4}
                          &                        & Crack Propagation      &     \citep{jiang2022anti,radeef2021characterisation,zhu2020digital,safavizadeh2017utilizing,zhang2018characterization,zhou2017cracking}     \\ \cline{2-4}
                          & Flexural fatigue test  & Crack Propagation      &    \citep{saride2019estimation}      \\
\hline
\multirow{2}{*}{Strength} & \multirow{2}{*}{IDT}   & Strain \& Displacement &    \citep{xing2020strain,jiao2020acoustic,xing2017research,tan2017design,xing2020particle,gorszczyk2019application}      \\ \cline{3-4}
                          &                        & Damage Evolution       &     \citep{al2022cracking,guo2020investigation,hasheminejad2019investigation,hasheminejad2018digital}     \\ \hline
Others                    & Pull-off adhesion test & Mechanistic Parameters    &    \citep{sedghi2021evaluating}      \\ \cline{2-4}
                          & \multirow{2}{*}{DTT}   & Strain \& Displacement &     \citep{roberto2021introducing,roberto2020effect,yan2018fracture}     \\ \cline{3-4}
                          &                        & Damage Evolution       &    \citep{roberto2020effect,preti2019effect,kumar2018evaluation}      \\ \cline{2-4}
                          & Freeze-thaw test       & Strain \& Displacement &     \citep{teguedi2017towards}     \\ \cline{2-4}
                          & Light healing test     & Strain \& Displacement &     \citep{wang2017experimental}    \\ \cline{2-4}
                          & 2T3C HCA     & Displacement &     \citep{attia20202t3c}    \\
\hline
\hline
\end{tabular}}
\caption{Applications of 2D-DIC in AC laboratory tests.}
\label{tab:2d_application}
\end{table}

\subsubsection{Direct Application of Strain/Displacement Fields}
The predominant focus in reviewed papers only involves the direct application of DIC-measured strain or displacement maps. These applications can be broadly categorized into two groups based on measurement scope: global, involving the entire displacement or strain fields of the specimen, and local, focused solely on the area of interest.

In global applications, researchers analyze chosen displacement or strain components (e.g., horizontal, vertical) at various time points during specimen loading to gain qualitative insights into material mechanical properties. This includes identifying high-strain zones on the specimen surface often associated with cracks or damage zones, which will be elaborated upon in the subsequent section dedicated to crack propagation measurement via DIC \citep{stewart2019fatigue,hasheminejad2019investigation,al2022cracking,al2015testing,birgisson2008determination}. Another significant application involves employing DIC as a validation tool for numerical models or displacement/strain sensors. DIC-measured displacement or strain fields serve as the "ground truth" to validate these models or sensors \citep{li2023study,radeef2022linear,hernandez2018micromechanical,behnke2021continuum,zobec2021application}. Additionally, DIC is employed as a measurement tool for deformation assessment, such as quantifying permanent deformations in repeated load uniaxial tests or cyclic uniaxial compression tests and determining vertical deformations in AC specimens during IDT \citep{behnke2021continuum,wen2002simple,gorszczyk2019application}.

In the context of local applications, researchers focus on specific regions within DIC-derived displacement and strain maps. A prominent use case involves the measurement of Crack Mouth Opening Displacement (CMOD) and the investigation of aggregate crushing near the load point. CMOD serves to characterize the displacement alteration perpendicular to the crack plane within a fatigued-notched region of a specimen subjected to fracture toughness testing \citep{anderson2017fracture}. CMOD measurements are taken along the loading axis or the specimen's surface, quantifying the difference between the initial and final crack openings. Traditionally, this is done with a physical clip-on displacement gauge attached to opposite sides of the crack mouth to generate a load-CMOD curve for fracture energy calculation. The application of DIC for measuring CMOD obviates the need for a physical gauge, replacing it with virtual measurement points positioned on opposite sides of the specimen \citep{hill2017inverse}. DIC-derived CMOD is a crucial metric for assessing AC's fracture characteristics and crack resistance. Specifically, fracture energy, representing the energy required to generate a unit area of crack, is a significant parameter in this context. It can be determined by calculating the area under the load-CMOD curve and subsequently dividing it by the ligament area, approximately equivalent to the product of the crack length and specimen thickness \citep{anderson2017fracture, hill2017inverse, hu2022use, al2015testing, zhu2019influence, li2010using, wagnoner2005disk}.

Another crucial local application involves the investigation of potential aggregate crushing near the load point. This is of paramount importance due to the significant energy dissipation associated with such crushing, potentially resulting in an overestimation of energy contributions to fracture or fatigue crack formation. Such overestimations are undesirable when employing energy-based parameters for material property comparisons. For example, \citet{doll2015evaluation} performed SCB tests, wherein strain fields in the vicinity of the loading point were visually compared at multiple time points throughout the loading procedure. Their observations revealed no substantial (i.e., order of magnitude) differences, confirming the absence of significant aggregate crushing. Similar investigation can be found in the work of \citet{yang2021feasibility}.

\subsubsection{Tracking Crack Propagation}
Crack propagation monitoring in fracture or fatigue tests is a significant application of DIC \citep{vanlanduit2009digital,zhao2022high,zhao2020situ}. This application can be categorized into four distinct approaches based on the underlying principles: visualization-based, empirical-mechanistic-based, mechanistic-based, and computer-vision-based.

\begin{figure}[!ht]
  \centering
  \includegraphics[width=0.8\textwidth]{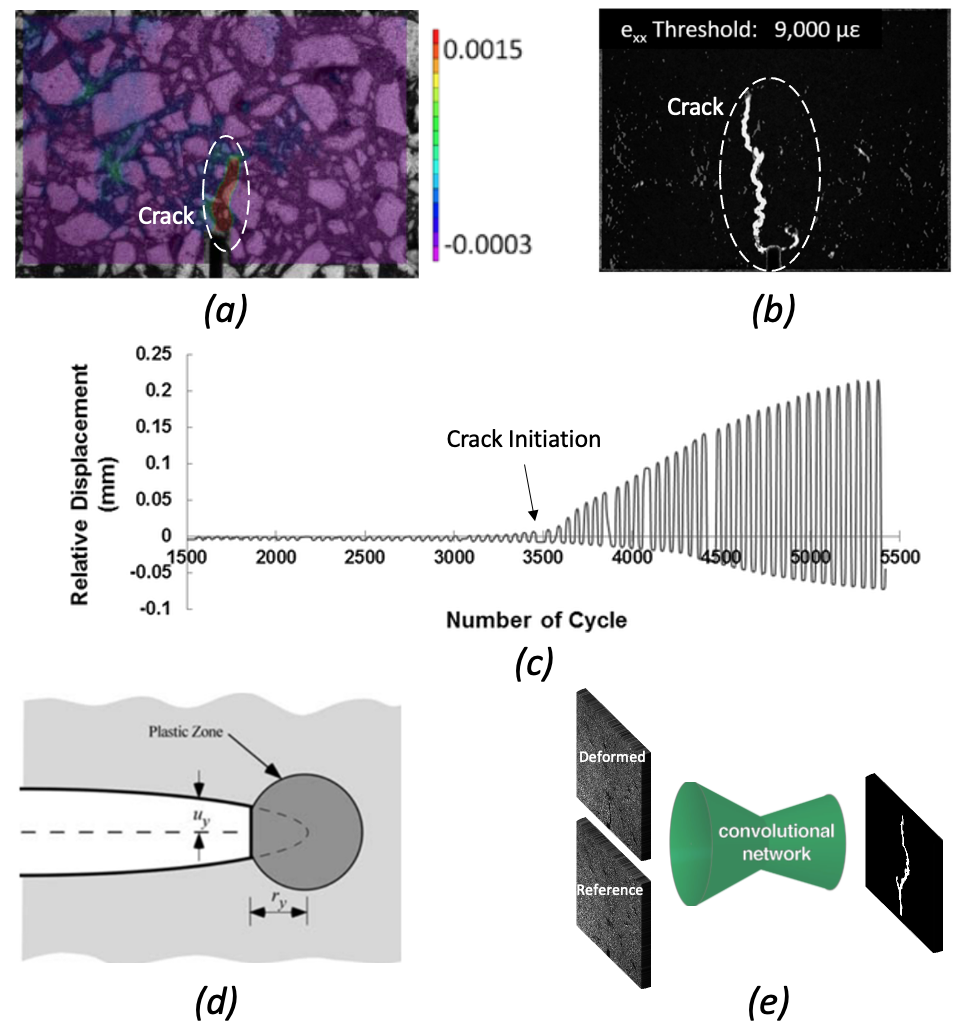}
  \caption{Tracking crack propagation (a) visualization-based approach; (b) strain thresholding approach; (c) deviation point assumption; (d) CTOD-based mechanistic approach; (e) CrackPropNet.}
  \label{fig:crack}
\end{figure}

In the visualization-based approach, researchers typically employ strain maps to discern crack tips and boundaries \citep{stewart2019fatigue,hasheminejad2019investigation,al2022cracking,al2015testing,birgisson2008determination}. High strains are typically visually detected and categorized as zones of damage or cracks (Figure \ref{fig:crack}\emph{(a)}). This method's advantage lies in its simplicity, enabling crack identification by a human expert without the need for subsequent processing of DIC-derived strain fields. However, its drawback is its subjective nature, leading to potential variations in crack propagation assessment among different individuals. Additionally, domain expertise is essential, as individuals lacking familiarity with the specific materials and testing procedures may misinterpret crack identifications.

In the empirical mechanistic approach, researchers often make use of empirical assumptions, such as strain thresholds or deviation points on relative displacement curves, to define the initiation of cracking. For example, \citet{safavizadeh2017dic} employed a threshold of 9,000 $\mu\epsilon$ for $e_{xx}$ and 6,000 $\mu\epsilon$ for $e_{yy}$ to detect vertical and interfacial cracks in double-layer grid-reinforced asphalt concrete notched beams under four-point bending fatigue loading (Figure \ref{fig:crack}\emph{(b)}). \citet{buttlar2014digital} assumed that the deviation point on a relative displacement versus number of cycles curve indicated failure at the layer interface in a double shear test (Figure \ref{fig:crack}\emph{(c)}). The advantage of this approach is its minimal post-processing requirements. However, it has the drawback of subjectivity and a reliance on domain-specific knowledge. Furthermore, the empirical assumptions are often challenging to validate or may remain unverifiable.

In the mechanics theory-based approach, researchers utilize fundamental mechanics theory to identify cracks. For example, \citet{zhu2023crack} proposed employing the critical crack tip opening displacement (CTOD) to define the onset of cleavage fracture (Figure \ref{fig:crack}\emph{(d)}). This proposed threshold holds physical significance and can be readily determined from DIC measurements. The advantage of this method lies in its reliance on well-established fundamental theory, reduced dependence on user inputs, and a higher likelihood of accurately representing actual cracks. However, it is more complex to implement compared to previous approaches and is less amenable to automation. It's important to note that Zhu's approach is specifically applicable to fracture tests and has been validated only for mode I fracture. Further research is recommended to develop methods suitable for fatigue tests and other fracture modes.

In the context of low-level computer vision, researchers treat strain or displacement maps as image data and apply classical computer vision techniques such as thresholding, edge detection, and blob extraction to identify cracks \citep{gehri2020automated,gehri2022refined,cheng2022evaluation}. This approach primarily relies on detecting abrupt changes in strain patterns, akin to sharp intensity changes in regular images. Its advantages include reduced subjectivity compared to prior methods and the potential for automation. However, a drawback lies in the necessity of using thresholds, which lack physical meaning and cannot be explicitly validated for their specific values. Another computer vision approach is based on optical flow, which characterizes the apparent motion patterns of speckle patterns \citep{horn1981determining}. \citet{zhu2023automated} introduced CrackPropNet, a deep neural network built upon optical flow principles (Figure \ref{fig:crack}\emph{(e)}). This network was trained on a comprehensive image dataset encompassing various types of crack behavior in AC. CrackPropNet takes a reference image and a deformed image as inputs, producing a probability map representing crack edges as its output. Notably, CrackPropNet achieves a high crack detection accuracy on AC specimen surfaces while maintaining a rapid processing speed of 26 frames per second.

It is important to emphasize that after measuring crack propagation, additional post-processing techniques can be employed to extract additional insights. For instance, one can construct an R-curve, which plots the crack growth resistance against the crack extension \citep{anderson2017fracture}. This R-curve-based approach acknowledges the fact that the fracture resistance of AC may not remain constant throughout the process of crack propagation \citep{radeef2023fracture,ghafari2015r}. Crack propagation measurement via DIC can also contribute to Paris's law, a prominent fatigue crack growth model that describes the connection between crack growth rate and stress intensity factor for asphalt concrete, as represented in Equation \ref{eqn:paris} \citep{zhang2018characterization,stewart2019fatigue,cheng2022evaluation}.

\begin{equation}
    \frac{da}{dN}=A(\Delta K)^n
\label{eqn:paris}
\end{equation}

where $a$ represents the crack length, $N$ is the loading cycle, and $A$ and $n$ denote the material parameters in Paris' law, while $\Delta K$ signifies the stress intensity factor amplitude \citep{paris1963critical}. $A$ and $n$ find wide application in various contexts, including the prediction of reflective cracking and the design of asphalt overlays \citep{elseifi2004simplified,zhou2010development}.

\subsubsection{Derivation of Mechanistic Parameters}
DIC-measured displacement or strain maps facilitate the determination of mechanistic parameters. These parameters can be categorized into two groups based on their processing complexity: direct and secondary. Direct parameters, such as CTOD, and fracture process zone (FPZ), require limited post-processing. In contrast, secondary parameters, including strain energy density, stress intensity factor (SIF) and J-integral, necessitate substantial post-processing. Further elaboration on these parameters is provided in subsequent sections.

\begin{figure}[!ht]
  \centering
  \includegraphics[width=0.9\textwidth]{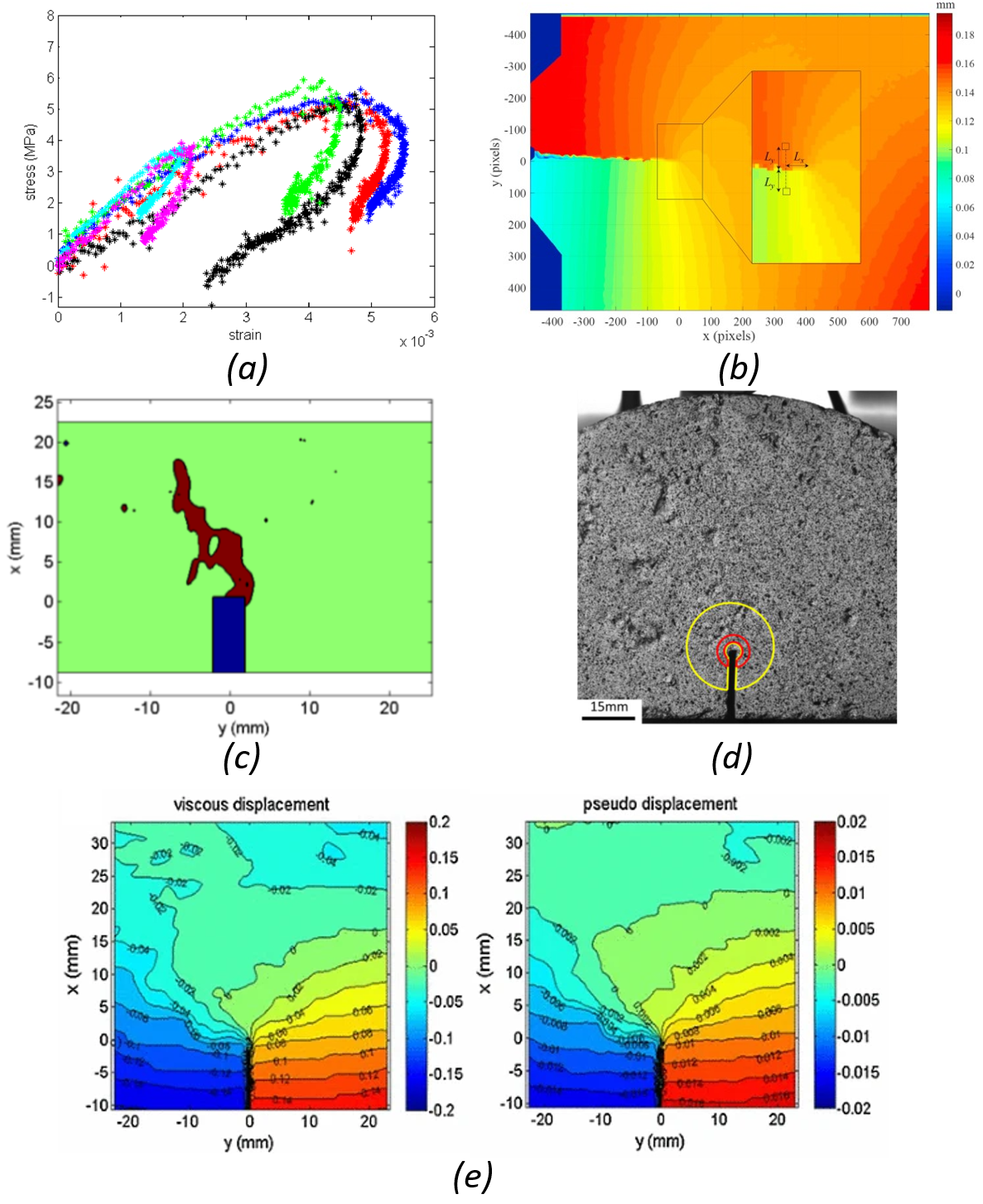}
  \caption{Combining mechanistic theory and DIC (a) stress-strain curve; (b) locate crack tip and select a pair of reference points for CTOD measurement; (c) eFPZ; (d) line J-integral around a notch; (e) viscous and pseudo displacement fields on a SCB specimen surface.}
  \label{fig:mecha}
\end{figure}

CTOD functions as a fracture mechanics parameter, particularly pertinent to elastic-plastic materials, signifying the material's resistance to crack propagation through strain at the crack tip \citep{kim2009discrete}. Physically measuring CTOD is often challenging; nevertheless, it can be deduced using a conventional plastic hinge model or a J-integral based model \citep{chen2014crack,anderson2017fracture,harrison1980cod,zhu2012review}. Moreover, accurately measuring CTOD using DIC is widely recognized as challenging due to the precise crack tip location and suitable reference point selection difficulties \citep{vasco2019characterisation}. \citet{zhu2023crack} adopted a method proposed by \citet{vasco2019characterisation} for measuring CTOD using DIC-measured displacement field data from a monotonic SCB test (Figure \ref{fig:mecha}\emph{(b)}). This approach entails two steps: first, locating the crack tip by plotting profiles of horizontal displacement perpendicular to the crack plane and determining the crack tip coordinates at the intersection of these profiles. Second, determining CTOD by defining a pair of reference points after locating the crack tip. By plotting relative displacements for various pairs of reference points, the appropriate reference point can be identified by identifying a stable plateau region, indicating the end of the strip-yield zone. It should be noted that this approach has only been applied to mode I fracture tests.

The DIC-measured strain field enables the determination of the stress-strain curve and the computation of strain energy density. The stress-strain curve characterizes AC's response to loading, offering information on its strength, stiffness, ductility, and failure thresholds. Strain energy denotes the energy stored within a deforming material, while strain energy density represents the energy stored per unit volume and corresponds to the integral area under the stress-strain curve, essentially encapsulating the area under the stress-strain curve. Strain energy density has been employed for fracture toughness prediction, fatigue damage characterization, and assessment of non-fracture-related energy dissipation \citep{doll2017investigation,aliha2019predicting,luo2013characterization}.

Asphalt concrete is a viscoelastic material, demonstrating both instantaneous elastic behavior and time-dependent viscous properties. This implies the presence of energy dissipation within the material when subjected to loading. Furthermore, AC's modulus is not constant, introducing complexity in deriving a stress field from the DIC-measured strain field. In accordance with established viscoelastic theory, the constitutive response of viscoelastic media can be expressed using the convolution Equation \ref{eqn:visco} \citep{christensen2012theory}.

\begin{equation}
    \sigma_{ij}(\xi)=\int_0^\xi E_{ijkl}(\xi-\xi')\frac{\partial \epsilon_{kl}(\xi')}{\partial \xi'}d\xi'
\label{eqn:visco}
\end{equation}

where $\sigma_{ij}$ and $\epsilon_{ij}$ represent stress and strain tensor components, $E_{ijkl}$ denotes the stiffness modulus components that depend on both time and temperature, and $\xi$ is a dimensionless time reduction parameter. $\xi$ is defined for thermo-rheologically simple materials as $\xi=t/a_T$, with $a_T$ representing a shift factor derived from the Williams–Landel–Ferry equation \citep{christensen1979rate,williams1955temperature}.

In a scenario where strain is directly determined as a function of applied load using DIC, the stress history was determined by evaluating the convolution integral described in Equation \ref{eqn:visco} at each load increment. To employ Equation \ref{eqn:visco}, it remains imperative to possess explicit knowledge of the temperature and time-dependent modulus $E$, typically expressed as a Prony series fit based on experimental data (Equation \ref{eqn:prony}).

\begin{equation}
    E(t)=E_e + \sum_{n=1}^N E_n e^{-\frac{t}{\rho_n}}
\label{eqn:prony}
\end{equation}

The equilibrium modulus, denoted as $E_e$, is a key parameter, while $E_n$ represents the Prony coefficients, and $\rho_n$ corresponds to the relaxation times.

Following this, one can plot the stress-strain curve (Figure \ref{fig:mecha}\emph{(a)}) and compute the strain energy density using Equation \ref{eqn:w}.

\begin{equation}
    W=\int_0^\epsilon \sigma_{ij}d\epsilon_{ij}
\label{eqn:w}
\end{equation}

The DIC-measured strain field facilitates the study of the FPZ, a region near the crack tip where material undergoes damage, even in the absence of complete cracking. This damage may manifest as microcracks, void formation, significant plastic deformation, or large-scale shearing (shear bands) \citep{hu1992fracture}. For AC, a strain-based approach inspired by \citet{wu2011experimental}'s work on concrete is considered effective. In this approach, the FPZ is defined as the zone where strains surpass a specified threshold value known as the tensile strain capacity, representing the maximum strain the material can endure before crack formation \citep{doll2017damage}. As precise tensile strain capacity values for different AC mixes are often unavailable, researchers commonly adopt a consistent, albeit arbitrary, threshold for comparative purposes. Consequently, instead of obtaining an absolute measurement of the FPZ extent for each mix, researchers calculate an estimated Fracture Process Zone (eFPZ), allowing for meaningful comparisons (Figure \ref{fig:mecha}\emph{(c)}). \citet{doll2017damage} have proposed thresholds of 3000 $\mu\epsilon$ at 25$^{\circ}$C and 1500 $\mu\epsilon$ at -12$^{\circ}$C.

Furthermore, DIC measurements allow for the computation of classical fracture parameters, including SIF and J-integral. In the context of linear elastic fracture mechanics, SIF serves as a predictive tool for assessing stress distributions near crack or notch tips induced by external loading or residual stresses. This parameter is contingent upon specimen geometry and loading conditions. For example, in mode I fracture, the onset of crack propagation is posited to arise when the applied stress intensity factor, denoted as $K_I$, surpasses a critical threshold known as fracture toughness ($K_{Ic}$) \citep{anderson2017fracture}. In experimental settings, DIC is employed to acquire displacement data, assuming the material behaves elastically. Subsequently, a least squares regression is executed using Equation \ref{eqn:320} to calculate $K_I$. However, it is important to note that asphalt material exhibits pronounced viscoelastic behavior, in contrast to the assumptions underlying Equation \ref{eqn:320} , which pertain to purely elastic materials. In cases of viscoelasticity, a similar approach can be adopted with the utilization of pseudo displacements as defined in Equation \ref{eqn:316} \citep{schapery1984correspondence} (Figure \ref{fig:mecha}\emph{(e)}). A least squares regression is then applied to these pseudo displacements using Equation \ref{eqn:320}, yielding $K_{IR}$ and accounting for rigid body motion. It is essential to acknowledge that the aforementioned procedure assumes a constant modulus for the asphalt material, which is erroneous in situations where the material exhibits high viscosity \citep{doll2015evaluation}.

\begin{equation}
    \begin{bmatrix} u_x \\ u_y \end{bmatrix} = \frac{K_I}{2\mu}\sqrt{\frac{r}{2\pi}}\begin{bmatrix} \cos(\frac{\theta}{2})[\kappa-1+2\sin^2(\frac{\theta}{2})] \\ \sin(\frac{\theta}{2})[\kappa+1-2\cos^2(\frac{\theta}{2})] \end{bmatrix}+\begin{bmatrix} u_{x0}-\theta_0 y \\ u_{y0}+\theta_0 x \end{bmatrix}
\label{eqn:320}
\end{equation}

Here, $\nu$ represents the Poisson ratio, $\mu$ stands for the shear modulus, $u_{x0}$ and $u_{y0}$ pertain to rigid translation, and $\theta_0$ represents rigid rotation. For plane strain, where $\kappa$ is defined as $\kappa=3-4\nu$. For plane stress, $\kappa$ is defined as $\kappa=\frac{3-\nu}{1+\nu}$.

\begin{equation}
    u_i^R=\frac{1}{E^R}\int_0^tE(t-t')\frac{\partial u_i}{\partial t'}dt'
\label{eqn:316}
\end{equation}

$E^R$ represents the reference modulus, which is typically selected as an arbitrary value, often taken as the instantaneous modulus denoted by $E_0$, while $E(t)$ signifies the relaxation function \citep{ozer2011development,tabatabaee2014establishing}.

The J-integral, unlike the SIF, is applicable to a broader range of material behaviors, including linear elastic, non-linear elastic, and plastic materials (under the condition of no unloading). It is valid for situations without body forces and in the context of 2D deformation fields. Theoretically, three primary methods exist for measuring the J-integral of a material. Firstly, the J-integral can be determined by analyzing DIC-measured displacement and strain fields (Figure \ref{fig:mecha}\emph{(d)}). Secondly, it can be obtained through tests involving multiple specimens with varied pre-crack (i.e., notch) lengths, while keeping other test parameters controlled \citep{wu2005fracture}. Thirdly, the J-integral can be measured using a single specimen by directly measuring the crack length \citep{huang2020determination}. Since this section focuses on the application of DIC, the details of the first approach will be presented. The J-integral exhibits path independence for hyperelastic materials when subjected to monotonic loading conditions. However, it loses this path-independence property if the material dissipates energy within the bulk. \citet{schapery1984correspondence} introduced an alternative formulation known as the pseudo J-integral, which is suitable for viscoelastic-homogeneous materials and is calculated along a line contour, which may pose numerical challenges when dealing with displacement derivatives from discrete data, such as those obtained from DIC measurements. An alternative expression (Equation \ref{eqn:jintegral}) is derived using Green's theorem (Divergence theorem) and is valid under specific conditions, including the absence of thermal strain, body forces, and traction along the crack faces \citep{nakamura1986analysis,doll2015evaluation}. It is crucial to highlight that the path independence of the J-integral relies on the assumption of material homogeneity, whereas AC exhibits heterogeneity.

\begin{equation}
    J=\int_A[\sigma_{ij}u_{j,i}-W\delta_{ij}]q_{1,i}dA
\label{eqn:jintegral}    
\end{equation}

where $\sigma_{ij}$ is the stress component, $u_{j,i}$ is the displacement, $W$ is the strain energy density of AC ($W=\int_0^\epsilon \sigma_{ij}d\epsilon_{ij}$). The expression involves the parameter $q_1$, which takes a value of 1 along the inner contour and 0 along the outer contour \citep{nakamura1986analysis}.

\section{3D-DIC}
The 2D-DIC method has been extensively studied in the pavement engineering community for characterizing asphalt concrete. However, it has specific requirements regarding specimen deformation, loading devices, and measuring systems. In cases where the surface of the specimen is not planar or experiences three-dimensional deformation following loading, the application of the 2D-DIC method is unfeasible \citep{pan2009two}. In response to these constraints, the 3D-DIC method, often referred to as stereo-DIC, has emerged as a solution \citep{luo1993accurate}. This technique entails the utilization of two synchronized cameras or a single camera equipped with custom light-splitting apparatuses grounded in binocular stereo vision principles. This section will discuss best practices for 3D-DIC imaging system setup, algorithms, and applications in asphalt concrete characterization.

\subsection{Imaging System}
\begin{figure}[!ht]
  \centering
  \includegraphics[width=0.8\textwidth]{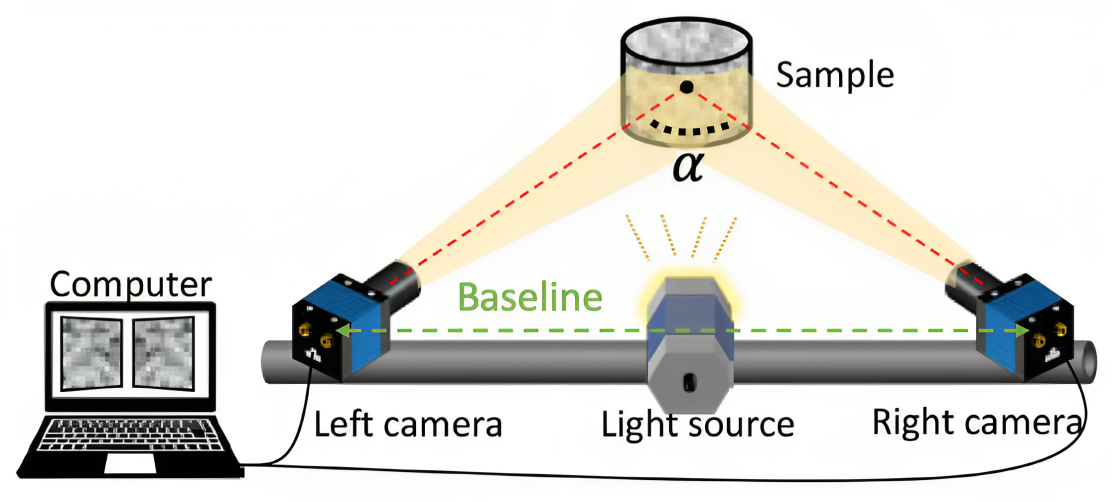}
  \caption{3D-DIC setup.}
  \label{fig:3d}
\end{figure}

A frequently employed 3D DIC setup comprises two synchronized cameras, an illumination setup, a computer, and a post-processing software (see Figure \ref{fig:3d}). Notably, the recommendations and guidelines discussed in a previous section regarding the setup of a 2D-DIC imaging system also apply to 3D-DIC. In addition, this section presents additional techniques to effectively set up a 3D-DIC imaging system. 

First, one essential requirement for 3D-DIC is the simultaneous acquisition of stereo images. However, achieving precise synchronization with minimal delay can be challenging. Generally, industrial cameras that support specific digital interface standards, such as CoaXPress, which has a built-in synchronization capability, are required to meet this demand \citep{reu2012stereo}.

Second, it is vital to determine the stereo angle ($\alpha$), which refers to the angular difference between the two camera views (Figure \ref{fig:3d}). Typically, narrower stereo angles (shorter baseline) enhance in-plane measurement accuracy but increase uncertainty in out-of-plane measurements. In scenarios emphasizing strain assessment, a narrower stereo angle is commonly favored. However, for improved out-of-plane results, it is recommended to use a larger stereo angle (longer baseline). When using a wide-angle lens, a stereo angle of at least 25$^\circ$ is advisable \citep{reu2013stereo}.

\subsection{Algorithm}
\subsubsection{Stereo Calibration}
To ensure accurate and high-quality 3D-DIC measurements, precise calibration of the two-camera unit used for simultaneous image capture, is crucial. Camera calibration furnishes essential parameters for triangulation, encompassing intrinsic details (such as center point, lens distortion coefficients, and individual camera focal lengths) and extrinsic factors (including translation and rotation between the dual cameras). A widely adopted and reliable method for calibration was proposed by \citet{zhang2000flexible}, which employs a 2D planar pattern. This technique is known for its high accuracy and ease of use, and it has become the standard method for calibration in most 3D-DIC techniques \citep{pan2018digital}.

The calibration process involves utilizing the captured stereo pairs of the 2D planar pattern. It is important to consider that the distortion-free projection of a 3D point $\widetilde{\mathbf{P}}_w=[x_1^w, x_2^w, x_3^w, 1]^T$ onto the camera sensor $\widetilde{\mathbf{p}}_c=[x_1^c, x_2^c, 1]^T$ can be represented by Equation \ref{eqn:proj}.

\begin{equation}
\begin{aligned}
    s^c\widetilde{\mathbf{p}}_c&=\mathbf{K}_c[\mathbf{R}_c|\mathbf{t}_c]\widetilde{\mathbf{P}}_w\\
    &=\begin{bmatrix}
        f_1^c & \gamma^c & c_1^c \\
        0 & f_2^c & c_2^c\\
        0 & 0 & 1\\
    \end{bmatrix}[\mathbf{R}_c|\mathbf{t}_c]\widetilde{\mathbf{P}}_w
\end{aligned}
\label{eqn:proj}
\end{equation}

Here, the subscript and superscript $c$ are used to indicate the camera indices. The symbol $s^c$ signifies a scaling factor, while $\mathbf{R}_c$ and $\mathbf{t}_c$ represent the rotation matrix and translation vector. These parameters facilitate the transformation from the world coordinate system to the camera coordinate system. $\mathbf{K}_c$ stands for the camera's intrinsic matrix, wherein $f_1^c$ and $f_2^c$ denote the focal lengths in pixels, $c_1^c$ and $c_2^c$ represent the pixel coordinates of the principal point (optical center), and $\gamma^c$ signifies the skew factor.

To achieve accurate calibration, it is essential to consider the non-linear optical distortion caused by the lens. A widely employed distortion model comprises radial and tangential distortions. These distortion coefficients are specific to each camera and are thus included as part of the camera's intrinsic parameters.

Subsequently, the stereo extrinsic parameters ($\mathbf{R}$ and $\mathbf{t}$) can be ascertained by evaluating the transformation equations connecting each camera, as described in Equations \ref{eqn:tran1} and \ref{eqn:tran2}.

\begin{equation}
    \mathbf{R} = \mathbf{R}_r\mathbf{R}_l^{-1}
\label{eqn:tran1}
\end{equation}

\begin{equation}
    \mathbf{t} = \mathbf{t}_r-\mathbf{R}_r\mathbf{R}_l^{-1}\mathbf{t}_l
\label{eqn:tran2}
\end{equation}

where subscripts $l$ and $r$ represent the left and right cameras, respectively. 

It is worth noting that stereo camera calibration can be conveniently performed using available tools such as Matlab's Stereo Camera Calibrator.

\subsubsection{Stereo Correlation}
The foundation of 3D-DIC lies in correlation algorithms to establish correspondence between points within left and right images. This correlation analysis comprises two key stages: stereo matching and temporal matching (or tracking). Stereo matching's primary goal is precise alignment of identical physical points present in the left and right camera images. Meanwhile, temporal matching tracks these identical points across successive images captured by the same camera at different instances or conditions. For temporal matching, the established subset-based 2D-DIC algorithm can be employed. Stereo matching is notably more intricate due to substantial perspective distortion between images captured by distinct cameras, rendering it the most challenging aspect of stereo vision measurement. To guarantee accurate and efficient 3D deformation measurements, crucial factors such as matching strategy, correlation algorithm, shape function, and initial estimation need careful consideration within the context of stereo matching \citep{pan2018digital}.

In stereo matching, the non-linear perspective projection frequently leads to substantial inaccuracies when employing first-order shape functions, especially when dealing with large subset sizes and considerable stereo angles. To tackle this issue, adopting $2^{nd}$-order shape functions in stereo matching is advisable, as it can enhance accuracy. For instance, \citet{gao2015high} introduced the inverse compositional Gauss-Newton (IC-GN$^2$) algorithm, employing $2^{nd}$-order shape functions. Concerning temporal matching, the IC-GN algorithm employing first-order shape functions is typically favored due to its greater computational efficiency.

Subset-based matching algorithms face challenges in estimating an initial guess when images from different cameras experience significant deformations caused by a large stereo angle. To overcome this limitation, feature-based matching techniques have been introduced. One such technique is the scale-invariant feature transformation (SIFT) algorithm, which allows for fast and accurate stereo matching in these scenarios \citep{lowe2004distinctive, zhou2012large}.

\begin{figure}[!ht]
  \centering
  \includegraphics[width=1\textwidth]{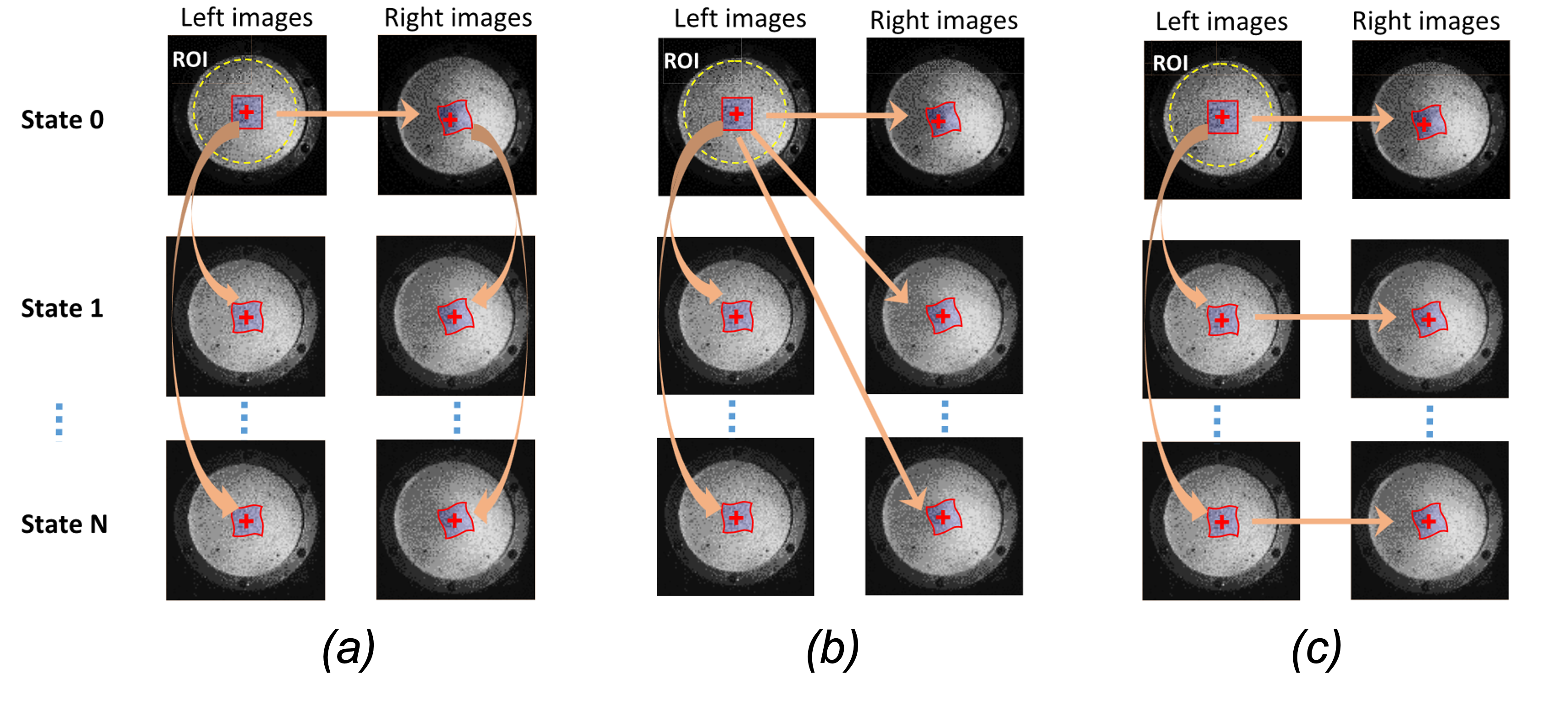}
  \caption{Strategies for stereo-correlations in 3D-DIC.}
  \label{fig:match}
\end{figure}

In the context of 3D-DIC, three prevalent matching strategies are employed. The initial approach (depicted in Figure \ref{fig:match}\emph{(a)}) entails matching the left reference image with the right reference image (for the initial state), and subsequently aligning all left and right deformed images to their corresponding initial states (temporal matching). The second strategy (Figure \ref{fig:match}\emph{(b)}) involves correlating all deformed left and right images with the left reference image. The third strategy (Figure \ref{fig:match}\emph{(c)}) compares the deformed left images to the left reference image via temporal matching, followed by matching each corresponding right image with the present left image through stereo matching. Amid these strategies, the first one, which conducts the computationally intensive stereo matching only once, is often deemed the most effective choice for practical measurements \citep{pan2018digital}.

\subsubsection{Reconstruction}
After the correlation process, each point in the image is now matched relative to the reference image. By incorporating the calibrated parameters of the stereo camera-unit, the classic triangulation method can be utilized to recover the 3D coordinates of measurement points. Within the domain of 3D-DIC, four prevalent techniques are utilized for 3D reconstruction: the least square method, optimal method, mid-point method, and geometrical optimal method \citep{wang2011error, kanatani2008triangulation, fooladgar2013geometrical, zhong2018three}. A comparative analysis by \citet{zhong2019comparative} established that the least square method stands out for its superior computational efficiency while maintaining measurement accuracy.

\begin{figure}[!ht]
  \centering
  \includegraphics[width=0.6\textwidth]{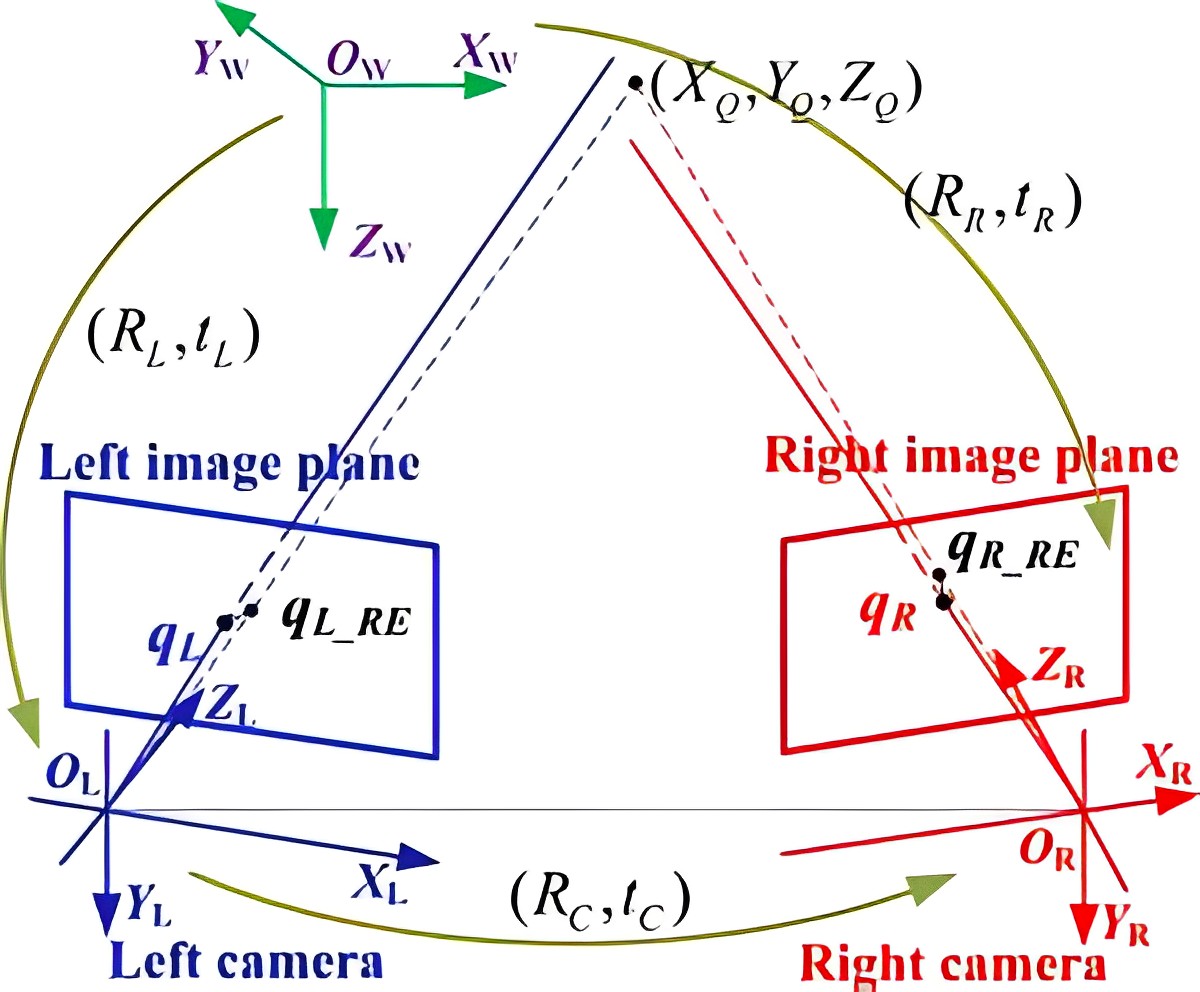}
  \caption{3D reconstruction based on triangulation.}
  \label{fig:reconstruct}
\end{figure}

The least square method computation is straightforward. Initially, the world coordinate system is established in alignment with the left camera coordinate system. Illustrated in Figure \ref{fig:reconstruct}, $\mathbf{R}_L$ corresponds to an identity matrix, and $\mathbf{t}_L$ represents a zero vector. $(\mathbf{R}_R, \mathbf{t}_R)$ is congruent to $(\mathbf{R}_C, \mathbf{t}_C)$, as obtained through stereo calibration. The point under computation is labeled as $Q$. Adopting the premise of a pinhole camera model, the correlation between the left image coordinates $(x_l, y_l)$ and the world coordinates $(X_Q, Y_Q, Z_Q)$ is expressed as elucidated in Equation \ref{eqn:re1}.

\begin{equation}
    \begin{bmatrix}
        x_l \\ y_l \\
    \end{bmatrix}=
    \begin{bmatrix}
        f_x^L & s_L & c_x^L \\
        0 & f_y^L & c_y^L \\
    \end{bmatrix}
    \begin{bmatrix}
        X_Q/Z_Q \\
        Y_Q/Z_Q \\
        1 \\
    \end{bmatrix}
\label{eqn:re1}
\end{equation}

Given, $\mathbf{R}_c = \begin{bmatrix}R_{11} & R_{12} & R_{13} \\R_{21} & R_{22} & R_{23} \\R_{31} & R_{32} & R_{33} \\\end{bmatrix}$ and $\mathbf{t}_c=\begin{bmatrix}t_x \\ t_y \\ t_z\end{bmatrix}$. The relationship between the right image coordinates $(x_r, y_r)$ and the world coordinates $(X_Q, Y_Q, Z_Q)$ can be described in Equation \ref{eqn:re2}.

\begin{equation}
    \begin{bmatrix}
        x_r \\ y_r \\
    \end{bmatrix}=
    \begin{bmatrix}
        f_x^R & s_R & c_x^R \\
        0 & f_y^R & c_y^R \\
    \end{bmatrix}
    \begin{bmatrix}
        \frac{R_{11}X_Q+R_{12}Y_Q+R_{13}Z_Q+t_x}{R_{31}X_Q+R_{32}Y_Q+R_{33}Z_Q+t_z} \\
        \frac{R_{21}X_Q+R_{22}Y_Q+R_{23}Z_Q+t_y}{R_{31}X_Q+R_{32}Y_Q+R_{33}Z_Q+t_z} \\
        1 \\
    \end{bmatrix}
\label{eqn:re2}
\end{equation}

By integrating Equations \ref{eqn:re1} and \ref{eqn:re2}, the world coordinates $(X_Q, Y_Q, Z_Q)$ can be reconstructed using Equation \ref{eqn:re3}.

\begin{equation}
    \begin{bmatrix}X_Q & Y_Q & Z_Q\end{bmatrix}^T = (\mathbf{M}^T\mathbf{M})^{-1}\mathbf{M}^T\mathbf{b}
\label{eqn:re3}
\end{equation}

$\mathbf{M}$ and $\mathbf{b}$ are given by Equations \ref{eqn:m} and \ref{eqn:b}, respectively. It is important to note that all the parameters in $\mathbf{M}$ and $\mathbf{b}$ were obtained during the stereo calibration process.

\begin{equation}
\resizebox{0.9\textwidth}{!}
{
    $\mathbf{M}=
    \begin{bmatrix}
        f_x^L & s_L & c_x^L-x_l \\
        0 & f_y^L & c_y^L-y_l \\
        R_{11}f_x^R+R_{21}s_R+R_{31}(c_x^R-x_r) & R_{12}f_x^R+R_{22}s_R+R_{32}(c_x^R-x_r) & 
        R_{13}f_x^R+R_{23}s_R+R_{33}(c_x^R-x_r) \\
        R_{21}f_y^R+R_{31}(c_y^R-y_r) &
        R_{22}f_y^R+R_{32}(c_y^R-y_r) &
        R_{23}f_y^R+R_{33}(c_y^R-y_r)
    \end{bmatrix}$
}
\label{eqn:m}
\end{equation}

\begin{equation}
    \mathbf{b}=
    \begin{bmatrix}
        0 \\
        0 \\
        -(t_xf_x^r+t_ys_R+t_z(c_x^R-x_r))\\
        -(t_yf_y^r+t_z(c_y^R-y_r))\\
    \end{bmatrix}
\label{eqn:b}
\end{equation}

\subsubsection{Compute of Displacements and Strains}
The previously acquired 3D coordinates serve as the basis for constructing complete displacement and strain maps. Displacements are individually computed for each point, while for each triangular element, strains are determined through the application of the Cosserat point element method \citep{solav2018multidic, solav2022duodic}. The vertices' positional vectors of each triangular element are employed to calculate the deformation gradient tensor $\mathbf{F}$. Utilizing $\mathbf{F}$, both the right and left Cauchy-Green deformation tensors ($\mathbf{C}=\mathbf{F}^T\mathbf{F}$ and $\mathbf{B}=\mathbf{F}\mathbf{F}^T$) are derived, alongside the Green-Lagrangian and Eulerian-Almansi strain tensors ($\mathbf{E}=0.5(\mathbf{C}-\mathbf{I})$ and $\mathbf{e}=0.5(\mathbf{I}-\mathbf{B}^{-1})$, respectively). By analyzing these tensors, principal components and directions are determined, leading to the derivation of principal stretches ($\lambda_i$) and strains ($E_i$ and $e_i$), as well as measures like equivalent (Von-Mises) strain, maximal shear strain, and area change.

\subsubsection{Software}
The effective implementation of algorithms is crucial for 3D-DIC analysis. Table \ref{tab:3d} presents a compilation of presently accessible non-commercial 3D-DIC software. It is essential to note that this list exclusively comprises software with supporting peer-reviewed research papers, and there may be other available choices.

\begin{table}[!ht]
\centering
\resizebox{\textwidth}{!}{\begin{tabular}{c|c|c|c|c|c|c|c}
\hline
\hline
Software & Authors  & First Release                & User Interface    & Open Source & Free & Language & Citations (Oct 2023) \\
\hline
DICe     & \citet{turner2015digital} & 2015      & Yes               & Yes         & Yes  & C++ & 12 \\
\hline
MultiDIC    & \citet{solav2018multidic} & 2018 & Yes & Yes         & Yes  & Matlab & 142 \\
\hline
DuoDIC   & \citet{solav2022duodic} & 2022 & Yes & Yes         & Yes  & Matlab & 4 \\
\hline
iCorrVision-3D   & \citet{nunes2022icorrvision} & 2022 & Yes & Yes         & Yes  & Python & 2 \\
\hline
\hline
\end{tabular}}
\caption{3D-DIC Software.}
\label{tab:3d}
\end{table}

\subsection{Applications}
The 3D-DIC technique offers advantages over the simpler 2D-DIC technique, as it does not make assumptions about the planar surface of the specimen or negligible out-of-plane displacement during testing. \citet{yuan2020full} investigated the out-of-plane deformation in both monotonic (fracture) and cyclic (fatigue) SCB tests using 3D-DIC. The results showed that in the fracture test, the out-of-plane displacement ranged from -0.45 mm to 0.45 mm, while in the fatigue test, it fluctuated between -1 mm and 0.95 mm. In a separate study, \citet{cheng2022evaluation} arrived at a similar conclusion. These findings indicate that the widely accepted assumption of negligible out-of-plane displacement in SCB tests may not be valid, further highlighting the advantages of 3D-DIC.

However, the adoption of the 3D-DIC technique in characterizing AC has been limited, with less than 5\% of journal publications since 2002 utilizing this technique. The first instance of such usage was reported in 2017 \citep{stewart2017comparison}. This limited adoption can be primarily attributed to the requirement of two synchronized cameras and a relatively complex camera calibration process \citep{zhu2023sift}. Table \ref{tab:3d_application} summarizes the applications of 3D-DIC in laboratory characterization of AC. The current applications of 3D-DIC closely resemble those of the 2D alternative, emphasizing the monitoring of displacement and strain map evolution during testing and the tracking of crack propagation. Future research may investigate the application of 3D-DIC in laboratory tests that are not amenable to 2D-DIC. For example, the use of Linear Variable Differential Transformers (LVDTs) is a common method for monitoring vertical displacement in dynamic modulus (E*) tests on cylindrical specimens. However, LVDTs require periodic calibration, labor-intensive installation, and extensive training. Additionally, they provide only a limited number of discrete measurement points on the specimen's surface \citep{tran2006evaluation}. Conversely, adopting 3D-DIC may allow for full-field displacement data acquisition, potentially eliminating the need for LVDTs. Furthermore, it is crucial to evaluate the validity of the assumption of negligible out-of-plane deformation in tests other than the SCB.

\begin{table}[!ht]
\centering
\resizebox{\textwidth}{!}{\begin{tabular}{c|c|c|c}
\hline
\hline
Type                      & Test                        & Applications           & Publications \\ \hline
\multirow{4}{*}{Fracture} & \multirow{2}{*}{SCB}        & Strain \& Displacement &  \citep{yang2023effect,cheng2022evaluation,wang2020cracking,yuan2020full,li2017crack}     \\ \cline{3-4}
                          &                             & Crack Propagation      & \citep{cheng2022evaluation}          \\ \cline{2-4}
                          & Double cantilever beam (DCB)                         & Mechanistic Parameters &  \citep{rajan2018traction}           \\ \hline
\multirow{3}{*}{Fatigue}  & 4PB                         & Strain \& Displacement    &     \citep{freire2021crack}         \\ \cline{3-4}
                          &                             & Crack Propagation      &     \citep{freire2021crack}         \\ \cline{2-4}
                          &    SCB     & Strain \& Displacement &  \citep{yuan2020full}                 \\ \cline{2-4}
\hline
\multirow{2}{*}{Strength} & \multirow{2}{*}{IDT}        & Strain \& Displacement &     \citep{stewart2019fatigue}         \\ \cline{3-4}
                          &                             & Damage Evolution       &  \citep{stewart2019fatigue}            \\ \cline{2-4}
\hline
\multirow{1}{*}{Others}   & Repeated load uniaxial test & Displacement    &   \citep{behnke2021continuum}           \\      
\hline
\hline
\end{tabular}}
\caption{Applications of 3D-DIC in AC laboratory tests.}
\label{tab:3d_application}
\end{table}

\section{Emerging DIC Techniques}

\subsection{Digital Volume Correlation}
2D-DIC and 3D-DIC are restricted to measuring surface displacements and strains. However, the heterogeneous properties of AC can lead to inconclusive results. This section discusses DVC, which enables displacement and strain mapping within the interior of loaded samples \citep{bay1999digital,roberts2014application,buljac2018digital}. Although the asphalt pavement engineering community has not yet adopted the DVC technique, the information presented here intends to promote future research in this field.

\begin{figure}[!ht]
  \centering
  \includegraphics[width=0.8\textwidth]{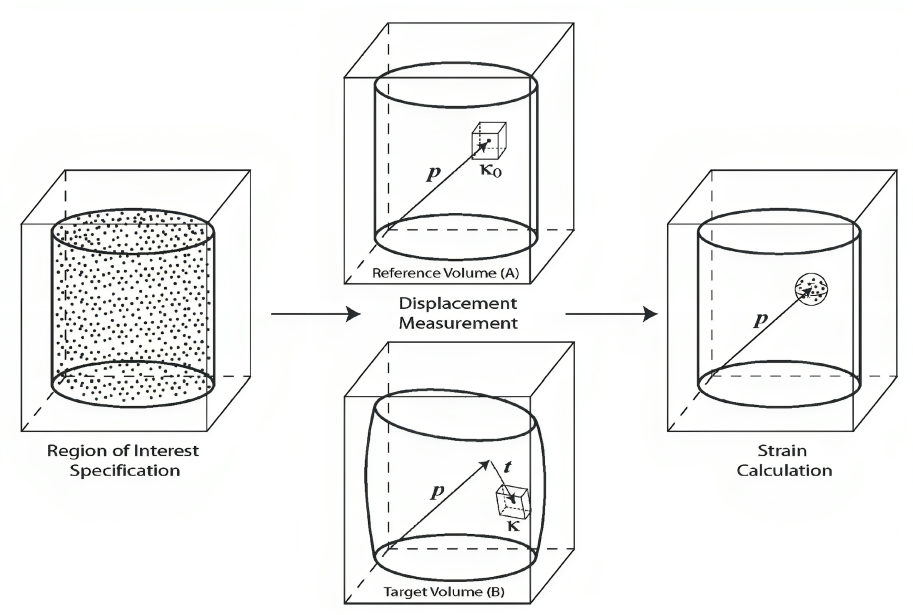}
  \caption{The overall digital volume correlation process.}
  \label{fig:dvc}
\end{figure}

The initial stage in applying DVC involves acquiring 3D images of unloaded and loaded specimens. X-ray computed tomography (CT) is the prevailing technique used for imaging, where a series of 2D X-ray images are used to generate 3D images of the specimens \citep{buljac2018digital}. It is important to mention that CT has been employed to investigate the internal structure of AC \citep{masad1999internal,song2005comprehensive,liu2014research}. Additionally, other imaging techniques, such as magnetic resonance imaging (MRI) and optical coherence tomography (OCT), can also be utilized \citep{benoit20093d,nahas20133d}.

As depicted in Figure \ref{fig:dvc}, the DVC process begins with the choice of an ROI encompassing the points requiring displacement determination.

Next in line is the estimation of displacement vectors at each measurement point. This is achieved through the correlation of a reference (unloaded) image volume with a target (deformed) image volume. Like 2D- and 3D-DIC methods, the calculation of displacement in DVC entails determining a combination of transformations (e.g., translation, shear, rotation) that minimizes a cost function, such as the sum-of-squares correlation coefficient (SSCC) or normalized cross-correlation coefficient (NCCC) cost function \citep{bay2008methods,pan2012internal,pan2014efficient}.

In the last phase, strains are assessed at all measurement sites by analyzing the deformation gradients within the neighboring vicinity. The strain tensor at each point $\mathbf{p}$ is calculated by fitting a $2^{nd}$-order Taylor series expansion of the displacement vector field using a group of nearby points through a least squares approach \citep{geers1996computing,pan2012internal}.


DVC presents promising solutions to common challenges in asphalt pavement engineering, including the validation of surface strain maps' representation of strain distribution across the entire specimen and the assessment of the correspondence between surface crack propagation measurements through DIC and actual three-dimensional crack propagation. Additionally, DVC demonstrates extensive prospective applications encompassing internal granular material movement tracking, internal strain quantification, crack initiation and fracture monitoring, computation of SIF along crack fronts, and analysis of fatigue crack closure effects, among others. Nevertheless, it is important to acknowledge that acquiring high-resolution 3D images capable of supporting DVC can be particularly challenging, especially given the complex and heterogeneous nature of AC \citep{du2019review}.

\subsection{Deep-Learning-Based DIC}
DIC is an iterative optimization procedure that requires substantial computational resources, resulting in extended calculation times. It also involves user inputs, including ROI, seed locations, subset size, and strain calculation window size, making it a non-automatic process. However, deep learning presents a solution to these challenges, enabling faster and fully automated DIC analysis, known as an end-to-end process.

To facilitate a comprehensive discussion on recent advancements in deep-learning-based DIC, it is crucial to introduce the concept of optical flow. Optical flow refers to the perceived displacement field derived from two views of a scene. It arises from the relative motion between specimens and the camera in the scene, encompassing movement and deformation \citep{fortun2015optical}. Notably, DIC and optical flow share common ground as both methodologies aim to determine the movement of pixels or features across a sequence of images. Deep learning techniques, including Convolutional Neural Networks (CNN), Recurrent Neural Networks (RNN), and Transformers, have been widely employed by researchers for optical flow estimation. Table \ref{tab:of} provides a concise overview of the most influential works in this area. Furthermore, the table presents the average endpoint error (AEPE) of these works on a well-known benchmark dataset, Sintel \citep{butler2012naturalistic}.

\begin{table}[!ht]
\centering
\resizebox{\textwidth}{!}{\begin{tabular}{c|c|c|c}
\hline
\hline
Method & Year  & Algorithm  & AEPE (in pixels) \\
\hline
FlowNet \citep{dosovitskiy2015flownet}   & 2015 &  Supervised; CNN  &  S (8.43); C (8.81)   \\
\hline
FlowNet2.0 \citep{ilg2017flownet}   & 2017 &  Supervised; CNN  &  6.02   \\
\hline
PWC-Net \citep{sun2018pwc}   & 2018 &  Supervised; CNN  &  5.04   \\
\hline
UnFlow \citep{meister2018unflow}   & 2018 &  Unsupervised; CNN  &  10.22   \\
\hline
RAFT \citep{teed2020raft}   & 2020 &  Supervised; RNN  &  2.86   \\
\hline
FlowFormer \citep{huang2022flowformer}   & 2022 &  Supervised; Transformer  &  2.09   \\
\hline
\hline
\end{tabular}}
\caption{Deep learning methods for optical flow estimation.}
\label{tab:of}
\end{table}

In 2021, \citet{boukhtache2021deep} introduced StrainNet, a deep-learning-based DIC method, for accurately determining in-plane subpixel displacement fields from pairs of reference and deformed speckle images. By fine-tuning FlowNet-S using a synthetic speckle dataset, StrainNet achieved a high level of accuracy, with a mean absolute error (MAE) of 0.0299 pixels. This accuracy is comparable to conventional 2D-DIC methods while also significantly reducing computation time. In a subsequent publication, the authors proposed a lightweight version called StrainNet-l \citep{boukhtache2023lightweight}. This version significantly reduced the number of parameters from 38.68 million to 0.67 million while maintaining a similar level of accuracy. StrainNet-l achieved an MAE of 0.0312 pixels, demonstrating that parameter reduction did not compromise its performance. \citet{wang2023dic} made further advancements in improving the accuracy of displacement field determination by training a CNN similar to U-Net architecture. They utilized a synthetic dataset generated through the application of the Hermite finite elements. The resulting network, named DIC-Net, achieved an impressive MAE of 0.0130 pixels, indicating a significant improvement in accuracy compared to previous methods. It is crucial to emphasize that the aforementioned networks are designed exclusively for retrieving displacement fields. To obtain strain maps, it is necessary to convolve the displacement fields with suitable derivative filters. However, \citet{yang2022deep} introduced an end-to-end network that directly measures strain maps using a FlowNet-S-like architecture. 

All the previously mentioned networks are applicable only to 2D-DIC, where a reference image and a deformed image are used as input. However, \citet{wang2022strainnet} developed a network called StrainNet-3D, designed explicitly for displacement retrieval from stereo images, similar to 3D-DIC. This approach incorporates an affine-transformation-based method for calculating disparities and a lightweight CNN for subpixel correlation. To ascertain three-dimensional displacement, the CNN-derived disparities and temporal optical flow are employed, guided by principles from stereo vision. Additionally, an optional refiner network can be employed to enhance the accuracy of the results further. StrainNet-3D achieved comparable accuracy to conventional 3D-DIC, with a mean absolute error (MAE) of 0.0146 pixels compared to 0.0110 pixels. Notably, StrainNet-3D exhibited improved efficiency as it does not require camera calibration and enables faster calculation.

Currently, deep-learning-based DIC methods utilizing RNN or transformer architectures have not been observed in the literature. However, based on the facts presented in Table \ref{tab:of}, these architectures have the potential to enhance the accuracy of deep-learning-based DIC approaches significantly.

Furthermore, it is worth noting that the pavement engineering community has not yet adopted these recent advancements in DIC using deep learning. Nevertheless, considering the advantages offered, such as improved computational efficiency, full automation, and elimination of user inputs compared to conventional DIC methods, it is recommended to initiate investigations into the viability of employing these deep-learning-based DIC methods for AC characterization.

\section{Flowchart of DIC Implementation for AC Characterization}
\begin{figure}[!ht]
  \centering
  \includegraphics[width=0.9\textwidth]{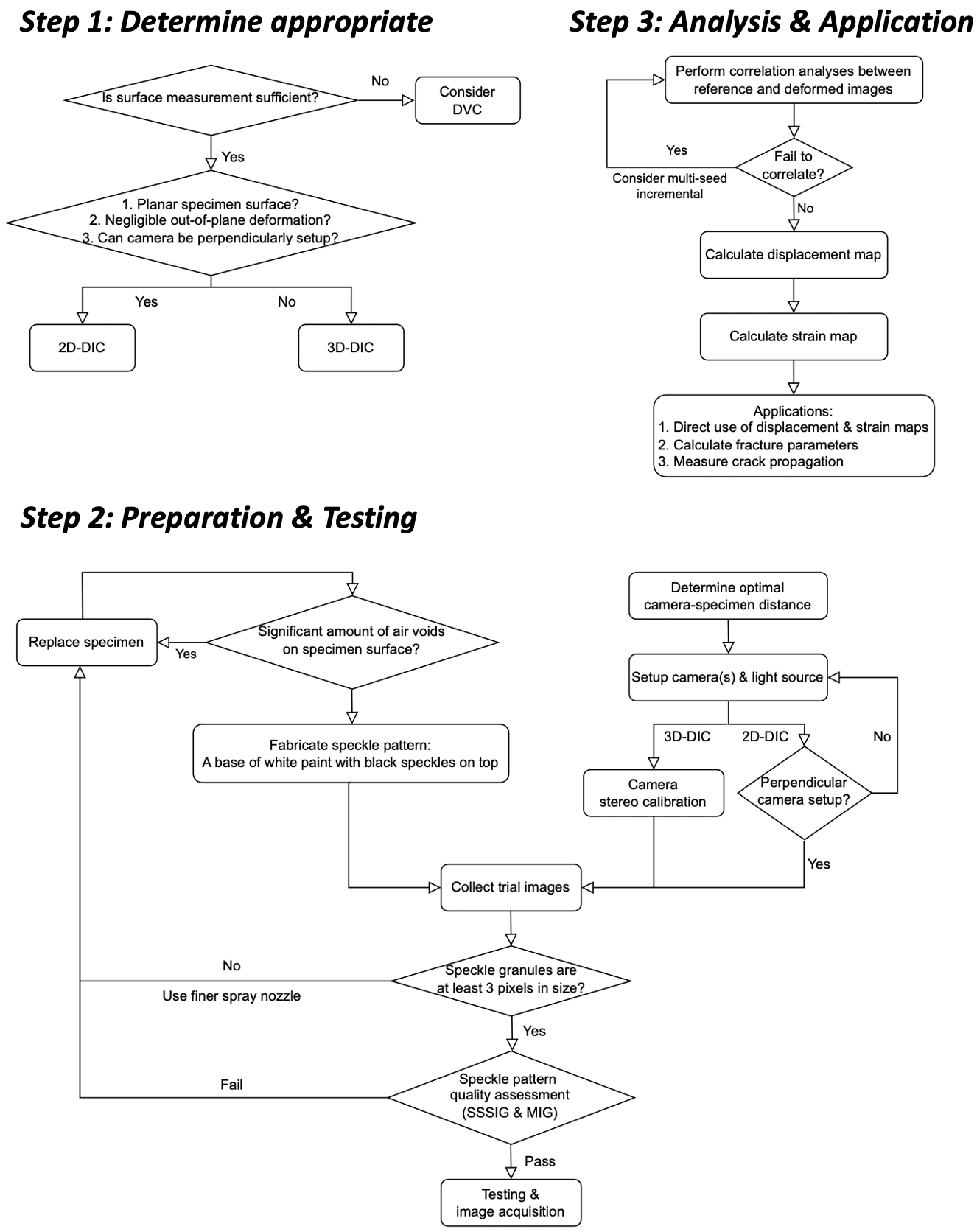}
  \caption{Flowchart for DIC implementation in AC characterization: synthesis of best practices from literature.}
  \label{fig:flowchart}
\end{figure}

The presented flowchart (Figure \ref{fig:flowchart}) serves as a comprehensive and reliable reference for implementing DIC in characterizing AC. It is based on a synthesis of best practices derived from the literature discussed in this paper. It is highly recommended to refer to the flowchart in conjunction with the detailed explanations provided in the respective sections of this paper for a thorough understanding of the implementation process.

\section{Summary and Recommendations for Future Research}
This article presents a comprehensive review of DIC as a critical tool for laboratory testing of AC. The focus is primarily on the widely used 2D-DIC and 3D-DIC techniques. The study thoroughly investigates best practices related to speckle pattern preparation, configuration of single-camera or dual-camera imaging systems, and meticulous execution of DIC analyses. Additionally, emerging DIC methodologies, such as DVC and deep-learning-based DIC, are introduced, highlighting their potential for future applications in pavement engineering. Lastly, a flowchart is provided as a comprehensive and reliable reference for implementing DIC in AC characterization.

The key takeaways are summarized as below: 

\begin{itemize}
    \item \textbf{\emph{Speckle Pattern Preparation}}. The optimal painted speckle pattern for AC specimen consists of black speckles applied onto a thin white basecoat. The speckle granules should ideally be 3-5 pixels or larger in size. SSSIG and MIG serve as effective indices for assessing the quality of the speckle pattern.
    \item \textbf{\emph{Imaging System Configuration}}. The optimal camera-specimen distance can be determined through mathematical calculations. To capture high-quality images with minimal noise, the three parameters of the exposure triangle, namely aperture, ISO, and shutter speed, need to be adjusted. Artificial lighting is often required to enhance the brightness of the scene. In 2D-DIC, it is advisable to employ a mechanical camera positioning tool to ensure parallel alignment between the camera CCD sensor and the object surface. For 3D-DIC, precise synchronization and stereo calibration of the dual-camera setup before the experiment are essential to achieve accurate measurements. A narrower stereo angle is preferable for in-plane measurement accuracy, while a larger angle is preferred for improved out-of-plane results.
    \item \textbf{\emph{Algorithm}}. In 2D-DIC, subset-based matching is performed between reference and deformed images. In parallel, 3D-DIC encompasses stereo matching, which strives to precisely align corresponding physical points in the images of the left and right cameras, and temporal matching, which monitors these identical points across successive images taken by the same camera under varying conditions or time frames. Open-source software options are readily accessible for both 2D- and 3D-DIC analyses.
    \item \textbf{\emph{Applications}}. DIC has found extensive application in fracture, fatigue, and strength tests. DIC has gained widespread utility in fracture, fatigue, and strength testing, categorizable into three main groups: direct application of DIC-generated displacement or strain maps, mechanistic parameter derivation, and tracking of crack propagation or damage evolution.
\end{itemize}

The followings are recommended for future research:
\begin{itemize}
    \item A scientific discourse exists concerning whether the natural texture of AC specimens aligns with prescribed criteria. In light of discrepant findings in existing research, it is advisable to conduct further investigations into the circumstances under which natural texture may be employed, considering factors such as mixture characteristics, imaging system configuration, and precision requirements.
    \item Most of the reviewed articles primarily employed DIC for displacement and strain measurement, with minimal post-processing.  Nevertheless, it is crucial to recognize that more meaningful and quantitative results, such as mechanistic parameters and precise crack propagation paths, can be derived through supplementary post-processing methods detailed in Section \ref{sec:2dapplications}.
    \item The prevailing approaches for tracking and quantifying crack propagation with DIC predominantly rely on visual or empirical methodologies. It is advisable to investigate the integration of fundamental mechanistic theories with DIC for the measurement of cracks in mode II fracture, mixed-mode fracture, and fatigue tests. Additionally, the combination of computer vision and fundamental mechanistic theories appears promising for achieving both high reliability and automation.
    \item Present methods for computing pseudo SIF and J-integral using strain fields obtained through DIC rely on the assumption of constant modulus and material homogeneity, respectively. Nevertheless, under high viscosity, the constant modulus assumption breaks down, and AC is inherently heterogeneous. Therefore, it is recommended to investigate approaches for computing pseudo SIF and J-integral under conditions where these assumptions do not hold.
    \item The utilization of 3D-DIC in AC characterization remains limited, accounting for less than 5\% of published articles in the field. Future research efforts could focus on implementing 3D-DIC in additional laboratory tests, particularly those where 2D-DIC is not feasible. Moreover, it is crucial to assess the validity of the assumption of negligible out-of-plane deformation in tests other than the SCB test and establish distinct guidelines for determining the appropriate use of 2D- or 3D-DIC in AC characterization.
    \item A deficiency in the application of DIC in large- or full-scale tests of AC has been observed. Cement concrete researchers have previously utilized DIC in such assessments. Notable challenges that must be addressed include optimizing the imaging system setup to minimize vibrations and achieve adequate spatial resolution, preparing specimens suitable for large-scale testing, and identifying the valuable insights attainable through DIC analysis.
    \item Both 2D-DIC and 3D-DIC are limited to surface displacement and strain measurements, which may yield inconclusive results due to the heterogeneous nature of AC. To overcome this limitation, it is recommended to investigate the potential of DVC as a tool for mapping displacement and strain within the interior of loaded AC samples.
    \item Deep-learning-based DIC methods have demonstrated enhanced computational efficiency, full automation, and reduced dependence on user inputs when compared to conventional DIC techniques. Therefore, it is recommended to explore the viability of utilizing these deep-learning-based DIC methods for characterizing AC.
\end{itemize}

\section*{Acknowledgements}
The authors extend their appreciation to the anonymous reviewers for their valuable feedback, which significantly enhanced the quality of this paper. This work received no external funding. Any commercial products mentioned in this paper do not represent endorsements from the authors.

\bibliographystyle{elsarticle-num-names} 
\bibliography{cas-refs}





\end{document}